\documentclass[5p,times,authoryear]{elsarticle}

\usepackage{amsmath}
\usepackage{amssymb}
\usepackage{graphicx}
\usepackage{subfigure}
\usepackage{booktabs}
\usepackage{epstopdf}
\usepackage{flushend}
 \usepackage[T1]{fontenc}
\usepackage{charter}
\usepackage{xcolor}
\definecolor{llmaccent}{HTML}{B3202C}
\definecolor{legacy}{HTML}{6B7C8C}
\usepackage{pgfplots}
\usepgfplotslibrary{groupplots}
\pgfplotsset{compat=1.16}
\usepackage{dblfloatfix}
\usepackage[english]{babel}
\usepackage{url}

\usepackage{microtype}                 
\journal{Quantitative Science Studies}

\begin{document}

\begin{frontmatter}

\title{Topical Phase Transitions in Artificial Intelligence
       Research: Large-Scale Evidence and an Early-Warning
       Signature for Emerging Topics}

\author[inst1]{Rasul Khanbayov\fnref{orcid1}}
\ead{rakh90100@hbku.edu.qa}
\fntext[orcid1]{ORCID: 0009-0005-9493-3390}

\author[inst1]{Hasan Kurban\corref{cor1}\fnref{orcid2}}
\ead{hkurban@hbku.edu.qa}
\fntext[orcid2]{ORCID: 0000-0003-3142-2866}


\cortext[cor1]{Corresponding author.}

\address[inst1]{College of Science and Engineering, Hamad Bin Khalifa University, Doha, Qatar.}

\begin{abstract}
Do research topics in artificial intelligence grow gradually, or do
they advance through abrupt, detectable jumps? Analyzing 80,814
accepted main-track papers from five premier AI conferences (ACL,
CVPR, ICLR, ICML, NeurIPS) spanning 2017 to 2025, we show major
AI topics advance through \emph{topical phase transitions}: remaining marginal for years, then surging across venues within one to three years. Large
language models became the dominant cross-venue topic by 2025, diffusion models rose with comparable abruptness, and
language-model methods crossed into computer vision via
vision-language models, whereas reinforcement learning compounded
smoothly, distinguishing genuine phase
transitions from ordinary growth. This structure is
our primary contribution: a large-scale, cross-venue characterization
of how AI research reorganizes. We then ask
whether a transition leaves a detectable footprint before it peaks.
We define an \emph{early-warning signature}, four
publication-dynamics criteria frozen on 2017--2021 data, and
evaluate it out of sample on 2023--2025 transitions, obtaining a
precision of 27\% and recall of 63\% against a 13.5\% base rate.
Applied to 2025 data, the signature flags reasoning and test-time
compute, agentic AI, multimodal LLMs, retrieval-augmented generation,
and world models as topics to monitor over 2026--2028. The source code is also publicly available on GitHub at \url{https://github.com/KurbanIntelligenceLab/ai-phase-transitions}.
\end{abstract}

\begin{keyword}
Topical phase transitions \sep
Early-warning signature \sep
Research trend forecasting \sep
Artificial intelligence \sep
Topic analysis \sep
Bibliometrics.
\end{keyword}

\end{frontmatter}


\section{Introduction}
\label{sec:intro}

As a branch of computer science, artificial intelligence (AI)
attempts to replicate human capacities for perception, reasoning,
and decision-making through machine learning and related methods
\citep{duan2019artificial}. Since its formal emergence as a
discipline in the mid-twentieth century, AI has evolved from
rule-based symbolic systems into the large-scale neural and
generative paradigms that now dominate research agendas worldwide.
Over the past decade, AI has been applied across domains including
healthcare \citep{triantafyllidis2019applications}, logistics
\citep{mahroof2019human}, business services, and disaster
prediction, reflecting its pervasive impact on social and economic
life.

At the same time, scientific publications in AI have experienced
breakthrough growth, reflecting the flourishing of the field
\citep{niu2016global}. This rapid expansion makes it
increasingly valuable to systematically track the research contents
and trends across AI's major publication venues. Scientific
publications, as the primary records of research activity, have long
been used to conduct historical analyses of a field. Such analyses
help scholars review past developments and speculate on future
directions. However, traditional narrative reviews are limited by
the number of publications they can cover and by the subjectivity
inherent in manual selection, which often restricts them to narrow
subfields \citep{haefner2021artificial}.

Various quantitative methods have therefore been applied to
characterize the historical development of AI research, from
publication and citation counts to co-occurrence and citation networks
and pre-defined topic categories. As we detail in
Section~\ref{sec:related}, these approaches summarize the state and
social structure of the field well but are limited in their ability to
characterize the content and temporal evolution of research topics
\citep{lee2018identifying}, and most have focused on journals rather
than the conferences where cutting-edge AI is predominantly published
\citep{yu2023discovering}. Understanding where the field is heading,
the primary concern of governments, industry, and academics, requires
methods aimed squarely at topical dynamics.

This study addresses these gaps and, beyond description, asks a
forward-looking question: can an impending paradigm shift be detected
from publication dynamics before it occurs? We conduct a
comprehensive longitudinal analysis of accepted main-track papers at
five major AI conferences (ACL, CVPR, ICLR, ICML, and NeurIPS)
spanning 2017 to 2025. We adopt a controlled-vocabulary keyword
extraction approach that produces directly interpretable topic labels
and supports systematic cross-venue comparison, and we aggregate
topic counts across temporal and venue dimensions to expose trends
that are invisible at the level of individual papers.

The central observation that motivates our predictive contribution is
that major AI topics do not advance smoothly. Instead, many follow
what we term a \emph{topical phase transition}: a topic remains
marginal for several years, then surges across multiple venues within
a one- to three-year window, a discontinuity qualitatively distinct
from steady compounding growth. We use the term in analogy to physical
phase transitions, where a system reorganizes abruptly past a critical
point, and we ground it empirically rather than metaphorically by
contrasting topics that undergo such transitions (large language
models, diffusion models) with one that grows smoothly over the same
period (reinforcement learning); this contrast is developed in
Section~\ref{sec:results}. This regularity makes early detection
plausible. Our contributions are threefold. First, and primarily, we
characterize the 2017--2025 AI research landscape and document its
topical phase-transition structure across topics and venues; this
large-scale, cross-venue characterization is the paper's main result.
Second, we operationalize this structure as an \emph{early-warning
signature}, four publication-dynamics criteria that tend to precede a
transition, and we evaluate it as a screening instrument by freezing
its thresholds on 2017--2021 data and measuring its historical hit
rate on the 2023--2025 transitions, reporting precision and recall
against the base rate. Third, we apply the signature to 2025 data to
flag candidate topics to monitor over 2026--2028, stated explicitly so
that they can be checked against future proceedings. We frame the
signature throughout as a calibrated screening heuristic with a
measured historical hit rate, not as a forecasting model with
guaranteed accuracy.

The remaining sections are organized as follows.
Section~\ref{sec:related} positions this study against prior
bibliometric and topic-modeling work on AI research.
Section~\ref{sec:data} introduces the methodology, covering data
collection and preprocessing, topic extraction and normalization, and
aggregation. Section~\ref{sec:results} documents the AI research
landscape and the phase-transition structure that motivates
prediction. Section~\ref{sec:signature} defines the pre-explosion
signature, validates it out of sample, and derives the 2026--2028
forecast. Section~\ref{sec:discussion} interprets the findings against
prior literature and acknowledges limitations.
Section~\ref{sec:conclusion} summarizes contributions and outlines
future work.

\section{Related Work}
\label{sec:related}

\subsection{Bibliometric Analysis of AI Research}

Quantitative studies of AI scholarship have characterized the field
along several dimensions. Publication and citation counts have been
used to identify influential authors, institutions, and countries
\citep{donthu2021conduct}, while co-occurrence networks of authors and
institutions expose collaborative structure \citep{niu2016global}, and
citation and co-citation analyses mine the knowledge structure of the
field \citep{small1973co}. These approaches summarize the
state and social organization of AI research well, but are limited in
their ability to characterize the content and temporal evolution of
research topics \citep{lee2018identifying}, which is the focus of the
present work.

\subsection{Topic-Modeling Studies of Research Trends}

A second line of work models topical content directly. Conventional
bibliometric studies have relied on keyword co-occurrence networks,
citation networks, and pre-defined topic categories
\citep{niu2016global, yu2019review}, which assign each paper to a
single topic from a fixed list and therefore struggle to discover
latent or emerging themes \citep{qian2020understanding}. Probabilistic
topic models, principally Latent Dirichlet Allocation (LDA), relax
this constraint. Most relevant is \citet{yu2023discovering}, who
applied LDA to 177,204 documents from 25 journals and 19 conferences
spanning 1990--2021, aggregating results across temporal, publication
source, and country-level perspectives. Abstract-based LDA excels at
discovering latent themes without prior vocabulary assumptions, whereas
the controlled-vocabulary keyword extraction used here produces labels
that are directly interpretable and lexically consistent across venues,
facilitating cross-venue comparison; prior work indicates the two
approaches are complementary, with LDA recovering a richer latent set
\citep{qian2020understanding} and keyword methods proving more robust
to disciplinary terminology variation \citep{lee2018identifying}.

\subsection{Positioning of This Study}

Relative to this body of work, the present study differs in three
respects. First, it concentrates exclusively on conference
publications from 2017 to 2025, the venues where cutting-edge AI is
predominantly published \citep{yu2023discovering}, yielding higher
temporal resolution during the period of most dramatic topical change
than journal-spanning analyses ending in 2021. Second, it analyzes
topic dynamics across five venues jointly, exposing inter-community
effects (such as the penetration of language-model methods into
computer vision) that are invisible to single-subfield studies. Third,
and most importantly, it moves beyond description: rather than only
characterizing past trends, it defines and validates a forward-looking
signature for detecting topical phase transitions before they occur, a
predictive contribution not present in prior descriptive bibliometric
work.

\section{Methodology}
\label{sec:data}

\begin{figure*}[htbp]
\centering
\includegraphics[width=\linewidth]{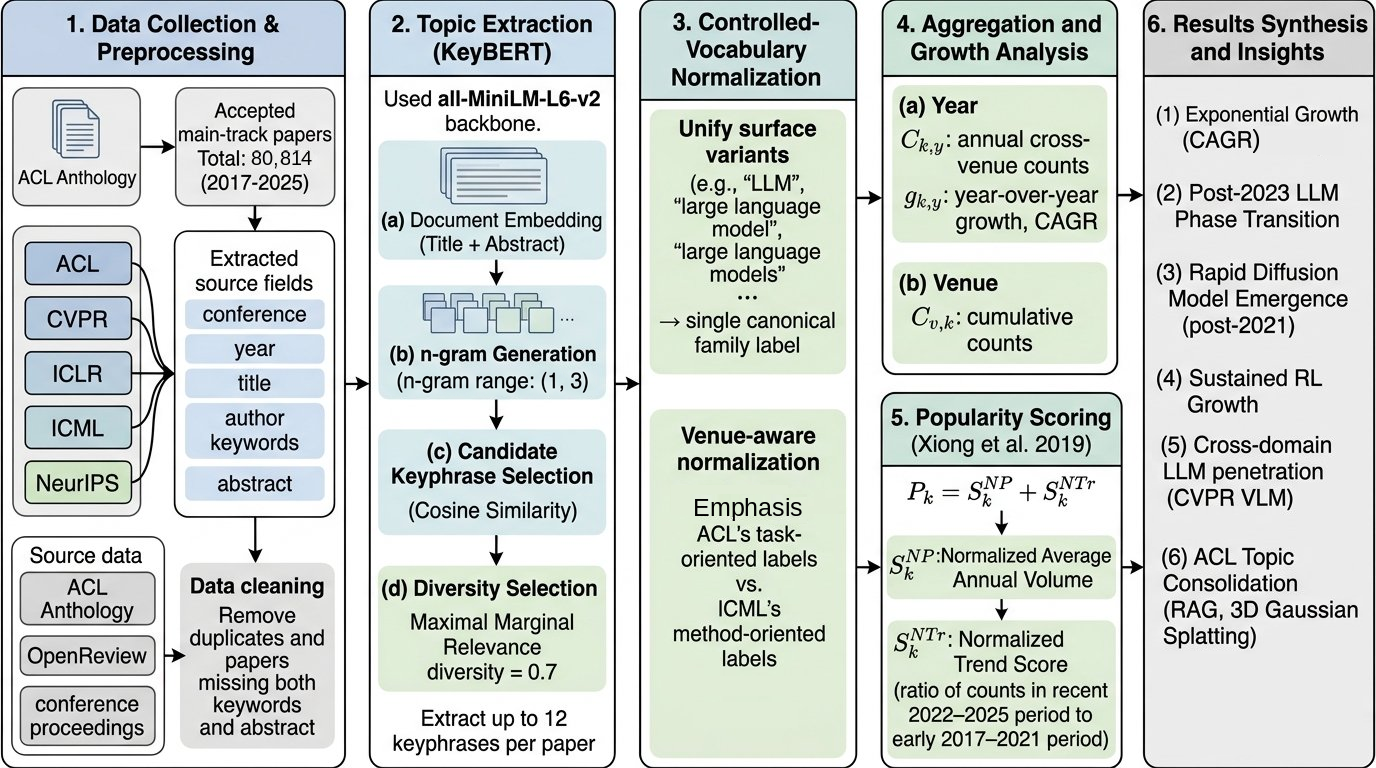}
\caption{Research pipeline: data collection and preprocessing (Stage~1),
KeyBERT-based topic extraction with maximal-marginal-relevance (MMR)
diversity selection (Stage~2),
controlled-vocabulary normalization (Stage~3), temporal and venue-level
aggregation with popularity scoring (Stage~4), results synthesis (Stage~5),
and forward prediction via the pre-explosion signature (Stage~6).}
\label{fig:framework}
\end{figure*}

The research framework is illustrated in Fig.~\ref{fig:framework}
and consists of three stages: (1)~data collection and preprocessing,
(2)~topic extraction and normalization, and (3)~topic aggregation
and analysis.

\subsection{Data Collection and Preprocessing}

In most bibliometric studies, corpora are assembled through keyword
searches of academic databases. For AI research, however,
keyword-based retrieval risks including large quantities of
peripheral content, since AI concepts appear across an exceptionally
wide range of disciplines. To assemble a more targeted and
authoritative dataset, this study selects five conferences widely
recognized as primary publication venues for NLP, computer vision,
and general machine learning research: ACL, CVPR, ICLR, ICML, and
NeurIPS. These venues are analogous in stature to the top-tier
conferences recommended by expert bodies in the field
\citep{qian2017citation}, ensuring a corpus of high-quality,
peer-reviewed research.

\begin{table}[htbp]
  \caption{Publication venues and accepted main-track paper counts
           by year (2017--2025). Years are shown as rows so the table
           fits a single column.}
  \label{tab:sources}
  \centering
  \footnotesize
  \setlength{\tabcolsep}{4pt}
  \begin{tabular*}{\columnwidth}{@{\extracolsep{\fill}}lrrrrrr@{}}
    \toprule
    Year & ACL & CVPR & ICLR & ICML & NeurIPS & Total \\
    \midrule
    2017 &   357 &   780 &   198 &   431 &   679 &  2,445 \\
    2018 &   673 &   977 &   337 &   618 & 1,009 &  3,614 \\
    2019 & 1,448 & 1,293 &   502 &   767 & 1,428 &  5,438 \\
    2020 & 1,263 & 1,461 &   687 & 1,080 & 1,898 &  6,389 \\
    2021 & 1,849 & 1,655 &   859 & 1,178 & 2,334 &  7,875 \\
    2022 & 1,651 & 2,070 & 1,094 & 1,231 & 2,687 &  8,733 \\
    2023 & 3,113 & 2,352 & 1,573 & 1,865 & 3,218 & 12,121 \\
    2024 & 2,755 & 2,711 & 2,260 & 2,638 & 4,035 & 14,399 \\
    2025 & 4,547 & 2,862 & 3,703 & 3,402 & 5,286 & 19,800 \\
    \midrule
    \textbf{Total} & \textbf{17,656} & \textbf{16,161} & \textbf{11,213}
      & \textbf{13,210} & \textbf{22,574} & \textbf{80,814} \\
    \bottomrule
  \end{tabular*}
\end{table}

For each conference, accepted main-track papers were collected for
the years 2017--2025. Metadata fields retained
for each paper include conference name, year, paper title, and
author-provided keywords or, where not available, the paper
abstract. After removing duplicate entries and papers lacking both
keywords and abstracts, the final dataset comprises approximately
80,814 papers. Table~\ref{tab:sources} summarizes the paper
counts per conference and year.

\subsection{Topic Extraction and Normalization}
\label{ssec:topic_extraction}

Topic labels were derived via a two-stage keyword extraction
pipeline applied to paper titles and, where available,
author-supplied keywords. In the first stage, candidate keywords
were scored using KeyBERT~\citep{grootendorst2020keybert}, a BERT-based
method that ranks candidate phrases by their cosine similarity to the
document embedding, with MMR applied to promote
lexical diversity.
In the second stage, candidates were normalized against a
conference-specific controlled vocabulary to reduce lexical
fragmentation. Closely related surface forms were unified into a
single canonical label; for example, \emph{``LLM''},
\emph{``large language model''}, and \emph{``large language
models''} were treated as a unified concept family for cross-venue
aggregation, while being retained as distinct labels for
fine-grained intra-venue analysis. This normalization step is
important because community terminology conventions differ
systematically across venues: ACL authors tend to use
task-oriented labels (e.g., \emph{``named entity recognition''}),
while ICML authors prefer method-oriented labels (e.g.,
\emph{``reinforcement learning''}).

\subsection{Topic Aggregation and Analysis}
\label{ssec:aggregation}

The static per-paper topic assignments produced by the extraction
pipeline cannot directly reveal the developmental trajectory of
research topics. To expose these dynamics, this study aggregates
topic counts from two complementary perspectives: time and
publication venue.

\subsubsection{Topic Trends Over Time}

For each topic $k$ and year $y$, the annual cross-venue paper
count is defined as:
\begin{equation}
    C_{k,y} = \sum_{v \in \mathcal{V}} c_{v,k,y},
    \label{eq:annual}
\end{equation}
where $c_{v,k,y}$ denotes the number of papers at venue $v$
assigned topic $k$ in year $y$, and
$\mathcal{V}=\{$ACL, CVPR, ICLR, ICML, NeurIPS$\}$.
This yields a time series
$[C_{k,2017}, C_{k,2018}, \ldots, C_{k,2025}]$ for each topic,
from which trends and inflection points can be identified.
Year-over-year growth rates are computed as:
\begin{equation}
    g_{k,y} = \frac{C_{k,y} - C_{k,y-1}}{C_{k,y-1}} \times 100\%,
    \label{eq:growth}
\end{equation}
and the compound annual growth rate (CAGR) over the full window as:
\begin{equation}
    \mathrm{CAGR}_k =
    \left(\frac{C_{k,T}}{C_{k,0}}\right)^{\!\!1/(T - t_0)} - 1.
    \label{eq:cagr}
\end{equation}

\subsubsection{Topic Distribution Over Publication Venue}

For each venue $v$ and topic $k$, the cumulative count is:
\begin{equation}
    C_{v,k} = \sum_{y} c_{v,k,y}.
    \label{eq:venue}
\end{equation}
These quantities enable both static comparison of topical
portfolios across venues and dynamic analysis of how each venue's
topical emphasis has shifted over time.

\subsubsection{Topic Popularity Scoring}

To comprehensively rank topics rather than relying solely on raw
counts or growth rates in isolation, this study adopts a composite
popularity score following the framework of
\citet{xiong2019analyzing}:
\begin{equation}
    P_k = S_k^{NP} + S_k^{NTr},
    \label{eq:popularity}
\end{equation}
where $S_k^{NP}$ is a normalized proportion score:
\begin{equation}
    S_k^{NP} =
    \frac{P_k^A - P_{\min}^A}{P_{\max}^A - P_{\min}^A},
    \label{eq:np}
\end{equation}
with $P_k^A$ denoting the average annual count of topic $k$; and
$S_k^{NTr}$ is a normalized trend score:
\begin{equation}
    S_k^{NTr} =
    \frac{S_k^{Tr} - S_{\min}^{Tr}}{S_{\max}^{Tr} - S_{\min}^{Tr}},
    \quad
    S_k^{Tr} =
    \frac{\displaystyle\sum_{y=2022}^{2025} C_{k,y}}
         {\displaystyle\sum_{y=2017}^{2021} C_{k,y}}.
    \label{eq:ntr}
\end{equation}
Here $S_k^{Tr}$ measures the ratio of recent-period counts to
early-period counts, capturing how much a topic has accelerated
relative to its prior baseline. Topics with high $P_k$ are
simultaneously prominent in volume and growing in trend; topics
with low $P_k$ may be large in absolute terms but declining in
relative share. The popularity ranking for all identified topics
is reported in Table~\ref{tab:popularity}.

\section{The AI Research Landscape and Its Phase-Transition Structure}
\label{sec:results}

This study obtains topic counts and distributions by implementing
the methodology described in Section~\ref{sec:data}. The findings
below characterize the AI research landscape from 2017 to 2025 and,
crucially, establish that its topics evolve through discrete phase
transitions. This phase-transition structure is what makes the
predictive signature of Section~\ref{sec:signature} possible: each
finding documents either a topic that surged abruptly (LLMs,
diffusion models, cross-domain vision-language models, abbreviated
VLMs) or one that did not (the steady
compounding of reinforcement learning), and the contrast between the
two is the empirical basis for early detection.

\subsection{Research Volume: Exponential Growth Across All Venues}
\label{ssec:growth}

Fig.~\ref{fig:growth} shows total accepted papers across all five
conferences from 2017 to 2025, together with a stacked
decomposition by venue. Total paper counts grew from approximately
2,445 in 2017 to nearly 19,800 in 2025, yielding a CAGR of
approximately 29.9\% by Eq.~\eqref{eq:cagr}. The trajectory
is super-linear: growth was moderate from 2017 to 2021, accelerated
noticeably from 2022 onward, and was most pronounced in the
2023--2025 window, when the single-year increment exceeded the
entire 2017 aggregate.

\begin{figure*}[tb]
\centering

\subfigure[Share of all papers by venue (all years combined).]{%
\begin{tikzpicture}[scale=1.0]
\def\R{2.0}
\foreach \s/\e/\c/\lab/\m in {
  90/-10.6/blue!60!black/NeurIPS 27.9\%/39.7,
  -10.6/-82.6/orange!85!black/CVPR 20.0\%/-46.6,
  -82.6/-161.2/green!50!black/ACL 21.8\%/-121.9,
  -161.2/-220.0/purple!70!red/ICML 16.3\%/-190.6,
  -220.0/-270.0/teal/ICLR 13.9\%/-245.0} {
  \fill[\c, fill opacity=0.82] (0,0) -- (\s:\R) arc (\s:\e:\R) -- cycle;
}
\node[font=\scriptsize, align=center] at (39.7:1.42*\R) {NeurIPS\\27.9\%};
\node[font=\scriptsize, align=center] at (-46.6:1.40*\R) {CVPR\\20.0\%};
\node[font=\scriptsize, align=center] at (-121.9:1.40*\R) {ACL\\21.8\%};
\node[font=\scriptsize, align=center] at (-190.6:1.46*\R) {ICML\\16.3\%};
\node[font=\scriptsize, align=center] at (-245.0:1.46*\R) {ICLR\\13.9\%};
\end{tikzpicture}
\label{fig:composition_pie}}
\hspace{0.05\linewidth}
\subfigure[Papers per year by venue (stacked bars).]{%
\begin{tikzpicture}
\begin{axis}[
  width=0.52\linewidth, height=5.4cm,
  ybar stacked, bar width=5.5pt, ymin=0,
  xtick={2017,2019,2021,2023,2025},
  scaled ticks=false, xticklabel style={/pgf/number format/1000 sep={}},
  yticklabel style={/pgf/number format/1000 sep={,}},
  xlabel={Year}, ylabel={Papers},
  tick label style={font=\scriptsize}, label style={font=\footnotesize},
  enlarge x limits=0.06,
  legend style={font=\scriptsize, at={(0.03,0.97)}, anchor=north west,
    legend columns=1, draw=gray!40, fill=white, fill opacity=0.85,
    text opacity=1, nodes={inner sep=1.5pt}},
  legend cell align=left]
\addplot[draw=none, fill=blue!60!black, fill opacity=0.82]
  coordinates {(2017,357)(2018,673)(2019,1448)(2020,1263)(2021,1849)(2022,1651)(2023,3113)(2024,2755)(2025,4547)};
\addlegendentry{ACL}
\addplot[draw=none, fill=orange!85!black, fill opacity=0.82]
  coordinates {(2017,780)(2018,977)(2019,1293)(2020,1461)(2021,1655)(2022,2070)(2023,2352)(2024,2711)(2025,2862)};
\addlegendentry{CVPR}
\addplot[draw=none, fill=green!50!black, fill opacity=0.82]
  coordinates {(2017,198)(2018,337)(2019,502)(2020,687)(2021,859)(2022,1094)(2023,1573)(2024,2260)(2025,3703)};
\addlegendentry{ICLR}
\addplot[draw=none, fill=purple!70!red, fill opacity=0.82]
  coordinates {(2017,431)(2018,618)(2019,767)(2020,1080)(2021,1178)(2022,1231)(2023,1865)(2024,2638)(2025,3402)};
\addlegendentry{ICML}
\addplot[draw=none, fill=teal, fill opacity=0.82]
  coordinates {(2017,679)(2018,1009)(2019,1428)(2020,1898)(2021,2334)(2022,2687)(2023,3218)(2024,4035)(2025,5286)};
\addlegendentry{NeurIPS}
\end{axis}
\end{tikzpicture}
\label{fig:composition_bar}}
\caption{Conference-level composition over 2017--2025.
(a)~Share of all papers by venue: NeurIPS contributes
27.9\%, followed by ACL (21.8\%), CVPR (20.0\%),
ICML (16.3\%), and ICLR (13.9\%). (b)~Papers per year by venue
(stacked bars); growth is broadly distributed across all five
venues, and the 2025 total reaches 19,800 papers.}
\label{fig:composition}
\end{figure*}
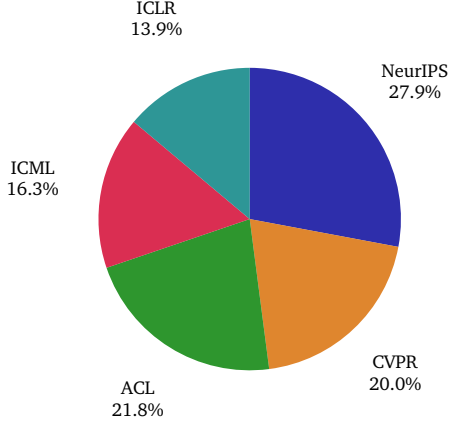
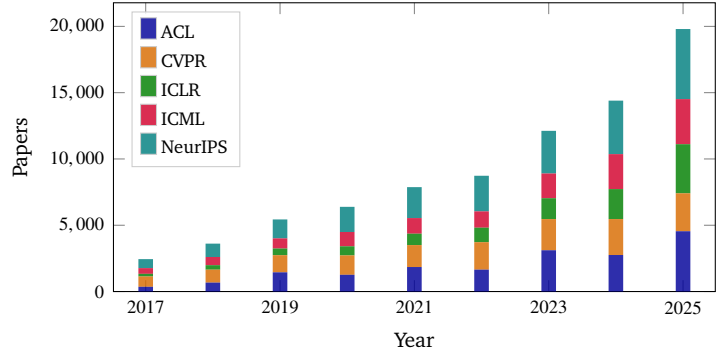

Fig.~\ref{fig:conference_comparison} disaggregates this growth by
conference for the two endpoint years, 2017 and 2025. NeurIPS was
the largest venue throughout, contributing approximately 5,286
papers in 2025 and 27.9\% of total cross-venue output across all
years (Fig.~\ref{fig:composition}a). ACL exhibited
the highest relative growth rate among the five venues, consistent
with the expanding scope of NLP research under the LLM paradigm.
CVPR is the third-largest venue overall (20.0\%) and grew steadily
in absolute terms. ICLR and ICML, which began
the period as smaller venues, expanded substantially, with ICML
growing nearly eightfold in annual output between 2017 and 2025. The stacked
bar chart in Fig.~\ref{fig:composition}b confirms that growth is
broadly distributed across all five venues rather than attributable
to any single conference, though NeurIPS and ACL collectively
account for a disproportionate share of the 2023--2025
acceleration.

These patterns are consistent with broader trends in AI investment,
researcher community expansion, and the increased accessibility of
publishing pipelines during this period. However, causal
attribution lies beyond the scope of the present analysis.

\begin{figure}[htbp]
\centering
\begin{tikzpicture}
\begin{axis}[
  name=top, width=0.92\columnwidth, height=4.2cm,
  ylabel={Total papers}, ymin=0, ymax=24000,
  xtick={2017,2019,2021,2023,2025},
  scaled ticks=false, xticklabel style={/pgf/number format/1000 sep={}},
  yticklabel style={/pgf/number format/1000 sep={,}},
  tick label style={font=\scriptsize}, label style={font=\footnotesize},
  grid=both, grid style={gray!18}, enlarge x limits=0.08,
  title style={font=\footnotesize}, title={Aggregate across five venues}]
\addplot[draw=blue!55!black, line width=1pt, mark=*, mark size=1.5pt,
  fill=blue!55!black, fill opacity=0.12]
  coordinates {(2017,2445)(2018,3614)(2019,5438)(2020,6389)(2021,7875)
               (2022,8733)(2023,12121)(2024,14399)(2025,19800)};
\node[font=\scriptsize, anchor=south] at (axis cs:2025,19800) {19,800};
\node[font=\scriptsize, anchor=south west] at (axis cs:2017,2445) {2,445};
\end{axis}
\begin{axis}[
  name=bot, at={(top.below south west)}, anchor=north west, yshift=-0.9cm,
  width=0.92\columnwidth, height=4.6cm,
  xlabel={Year}, ylabel={Papers}, ymin=0,
  xtick={2017,2018,2019,2020,2021,2022,2023,2024,2025},
  scaled ticks=false, xticklabel style={/pgf/number format/1000 sep={}},
  yticklabel style={/pgf/number format/1000 sep={,}},
  tick label style={font=\scriptsize}, label style={font=\footnotesize},
  grid=both, grid style={gray!18}, enlarge x limits=0.08,
  stack plots=y, area style,
  legend style={font=\scriptsize, at={(0.5,-0.42)}, anchor=north,
    legend columns=5, draw=gray!40,
    /tikz/every even column/.append style={column sep=6pt}},
  legend cell align=left,
  title style={font=\footnotesize}, title={Stacked volume by venue}]
\addplot[draw=none, fill=blue!60!black, fill opacity=0.85]
  coordinates {(2017,357)(2018,673)(2019,1448)(2020,1263)(2021,1849)
               (2022,1651)(2023,3113)(2024,2755)(2025,4547)} \closedcycle;
\addlegendentry{ACL}
\addplot[draw=none, fill=orange!85!black, fill opacity=0.85]
  coordinates {(2017,780)(2018,977)(2019,1293)(2020,1461)(2021,1655)
               (2022,2070)(2023,2352)(2024,2711)(2025,2862)} \closedcycle;
\addlegendentry{CVPR}
\addplot[draw=none, fill=green!50!black, fill opacity=0.85]
  coordinates {(2017,198)(2018,337)(2019,502)(2020,687)(2021,859)
               (2022,1094)(2023,1573)(2024,2260)(2025,3703)} \closedcycle;
\addlegendentry{ICLR}
\addplot[draw=none, fill=purple!70!red, fill opacity=0.85]
  coordinates {(2017,431)(2018,618)(2019,767)(2020,1080)(2021,1178)
               (2022,1231)(2023,1865)(2024,2638)(2025,3402)} \closedcycle;
\addlegendentry{ICML}
\addplot[draw=none, fill=teal, fill opacity=0.85]
  coordinates {(2017,679)(2018,1009)(2019,1428)(2020,1898)(2021,2334)
               (2022,2687)(2023,3218)(2024,4035)(2025,5286)} \closedcycle;
\addlegendentry{NeurIPS}
\end{axis}
\end{tikzpicture}
\caption{Exponential growth in total accepted papers across ACL,
CVPR, ICLR, ICML, and NeurIPS (2017--2025). Top: aggregate paper
count as a shaded line chart, growing from $\approx$2,445 in 2017
to 19,800 in 2025. Bottom: stacked area chart by venue, summing to
the same annual totals.}
\label{fig:growth}
\end{figure}
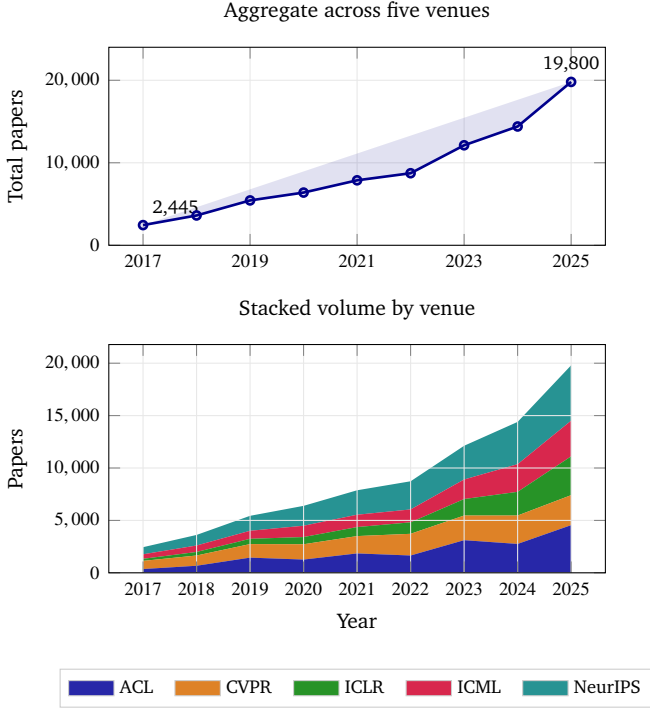

\begin{figure}[htbp]
\centering
\begin{tikzpicture}
\begin{axis}[
  width=0.92\columnwidth, height=5.2cm,
  ybar, bar width=7pt, ymin=0,
  symbolic x coords={ACL,CVPR,ICLR,ICML,NeurIPS},
  xtick=data, enlarge x limits=0.12,
  ylabel={Number of papers},
  yticklabel style={/pgf/number format/1000 sep={,}},
  tick label style={font=\footnotesize}, label style={font=\footnotesize},
  grid=both, grid style={gray!18},
  legend style={font=\footnotesize, at={(0.5,0.98)}, anchor=north,
    legend columns=-1, draw=gray!40,
    /tikz/every even column/.append style={column sep=10pt}},
  legend cell align=left,
  nodes near coords style={font=\tiny, /pgf/number format/1000 sep={,}}]
\addplot[draw=blue!55!black, fill=blue!55!black, fill opacity=0.8]
  coordinates {(ACL,357)(CVPR,780)(ICLR,198)(ICML,431)(NeurIPS,679)};
\addlegendentry{2017}
\addplot[draw=orange!85!black, fill=orange!85!black, fill opacity=0.85]
  coordinates {(ACL,4547)(CVPR,2862)(ICLR,3703)(ICML,3402)(NeurIPS,5286)};
\addlegendentry{2025}
\end{axis}
\end{tikzpicture}
\caption{Research volume by conference, 2017 versus 2025 (paper
counts). Each pair of bars contrasts a venue's output in the first and
last years of the study window. NeurIPS leads in 2025
($\approx$5,286 papers); ICLR shows the largest relative growth,
expanding roughly nineteenfold from its 2017 baseline of 198 papers.}
\label{fig:conference_comparison}
\end{figure}
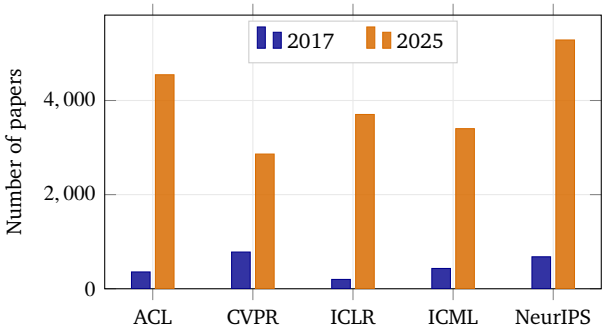

\subsection{The LLM Revolution: A Post-2023 Phase Transition}
\label{ssec:llm}

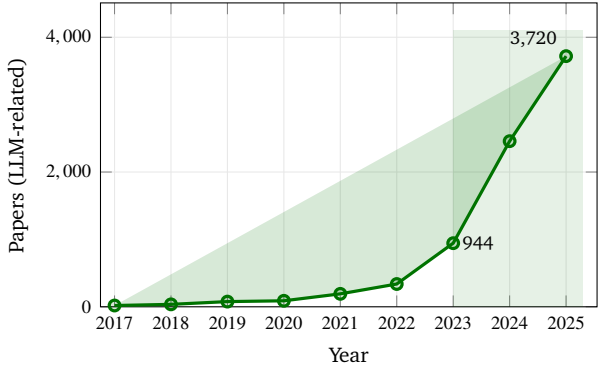
\begin{figure}[htbp]
\centering
\begin{tikzpicture}
\begin{axis}[
  width=0.92\columnwidth, height=5.6cm,
  xlabel={Year}, ylabel={Papers (LLM-related)}, ymin=0,
  xtick={2017,2018,2019,2020,2021,2022,2023,2024,2025},
  scaled ticks=false, xticklabel style={/pgf/number format/1000 sep={}},
  yticklabel style={/pgf/number format/1000 sep={,}},
  tick label style={font=\scriptsize}, label style={font=\footnotesize},
  grid=both, grid style={gray!18}, enlarge x limits=0.03]
\addplot[draw=none, fill=green!45!black, fill opacity=0.10, forget plot]
  coordinates {(2023,0)(2023,4100)(2025.3,4100)(2025.3,0)} \closedcycle;
\addplot[draw=green!45!black, line width=1.2pt, mark=*, mark size=2pt,
  fill=green!45!black, fill opacity=0.16]
  coordinates {(2017,19)(2018,36)(2019,77)(2020,89)(2021,192)
               (2022,338)(2023,944)(2024,2458)(2025,3720)};
\node[font=\scriptsize, anchor=south east] at (axis cs:2025,3720) {3,720};
\node[font=\scriptsize, anchor=west] at (axis cs:2023,944) {944};
\node[font=\scriptsize, green!35!black, anchor=south west]
  at (axis cs:2023.05,5400) {post-2023};
\end{axis}
\end{tikzpicture}
\caption{The LLM revolution: papers whose KeyBERT-extracted keywords
contain the phrase \emph{language model} (2017--2025). Counts grow
from under 340 per year before 2022 to 944 in 2023 and 3,720 in
2025. The shaded post-2023 region demarcates the phase transition.}
\label{fig:llm_revolution}
\end{figure}

Fig.~\ref{fig:llm_revolution} presents LLM-related paper counts
aggregated across all five venues from 2017 to 2025. The resulting
trajectory exhibits a qualitative discontinuity inconsistent with
gradual paradigm evolution: papers containing a language-model
keyword numbered fewer than 340 per year across all venues combined
through 2022, rose to 944 in 2023, and reached 3,720 by
2025, a nearly fourfold increase in two years. Applying
Eq.~\eqref{eq:ntr}, the trend ratio $S_k^{Tr}$ for
\emph{``large language models''} is $\approx$300, the highest in the
dataset by a substantial margin.

\begin{figure}[t]
\centering
\begin{tikzpicture}
\begin{axis}[
  width=0.8\columnwidth, height=5.0cm,
  xbar, bar width=14pt,
  xmin=0, xmax=1080,
  y dir=reverse,
  symbolic y coords={Large language models,Neural machine translation,Named entity recognition},
  ytick={Large language models,Neural machine translation,Named entity recognition},
  yticklabels={Large language models, Neural machine translation, Named entity recognition},
  enlarge y limits=0.6,
  axis lines=left,
  y axis line style={draw=none}, ytick style={draw=none},
  x axis line style={draw=none}, xtick style={draw=gray!45},
  xtick distance=300,
  xlabel={Cumulative papers, 2017 to 2025},
  xlabel style={font=\footnotesize, color=gray!55!black},
  xticklabel style={font=\scriptsize, color=gray!55!black,
                    /pgf/number format/1000 sep={,}},
  yticklabel style={font=\footnotesize},
  xmajorgrids=true, grid style={gray!18, line width=0.3pt},
  tick align=outside, clip=false,
  point meta=explicit symbolic,
  nodes near coords,
  nodes near coords style={font=\scriptsize, anchor=west, xshift=3pt, text=black},
]
\addplot[xbar, bar shift=0pt, draw=none, fill=llmaccent]
  coordinates {(892,Large language models) [892]};
\addplot[xbar, bar shift=0pt, draw=none, fill=legacy]
  coordinates {(262,Neural machine translation)
               [{262\quad\textcolor{gray!55!black}{$3.4\times$ smaller}}]};
\addplot[xbar, bar shift=0pt, draw=none, fill=legacy]
  coordinates {(183,Named entity recognition)
               [{183\quad\textcolor{gray!55!black}{$4.9\times$ smaller}}]};
\end{axis}
\end{tikzpicture}
\caption{At ACL, large language models dwarf the traditional NLP tasks
they have displaced. Cumulative accepted papers, 2017 to 2025: large
language models reach 892, against 262 for neural machine translation
and 183 for named entity recognition, factors of $3.4\times$ and
$4.9\times$ respectively. Red marks the emergent paradigm; grey marks
the legacy topics. The full top-25 ACL trajectory appears in
Fig.~\ref{fig:top25_acl} (Appendix~\ref{app:pervenue}).}
\label{fig:acl_combined}
\end{figure}
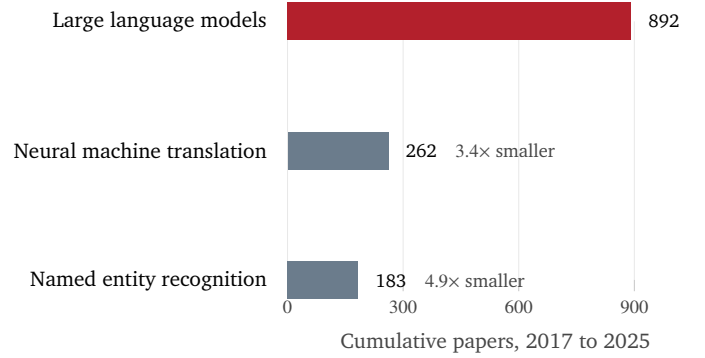

The venue-level breakdown makes the implications of this transition
concrete. The full top-25 ACL trajectory
(Fig.~\ref{fig:top25_acl}, \ref{app:pervenue}) shows that
\emph{``large language models''} diverges sharply
from all competing topics after 2022 and reaches 447
papers at ACL alone in 2025, nearly 90 times the count of 5
recorded in 2022. Fig.~\ref{fig:acl_combined} compares the
cumulative output of the leading LLM topic with traditional NLP
task categories. Across all years combined, \emph{``large language
models''} (892 papers) exceeds \emph{neural machine translation}
(262) by a factor of 3.4$\times$ and \emph{named entity
recognition} (183) by a factor of 4.9$\times$.
Neural machine translation, once a
defining research thread at ACL with a peak of 64
papers in 2019, has been effectively flat since 2022, while LLM
counts grew by over an order of magnitude in the same window.

\begin{figure}[tb]
\centering
\begin{tikzpicture}
\begin{axis}[
  width=0.70\columnwidth, height=10cm,
  enlargelimits=false, axis on top,
  colormap={ylord}{color=(yellow!15) color=(orange) color=(red!80!black)},
  colorbar horizontal,
  colorbar style={font=\scriptsize, xlabel={Papers},
    xlabel style={font=\scriptsize}, at={(0.5,-0.14)}, anchor=north,
    width=0.74*\pgfkeysvalueof{/pgfplots/parent axis width}},
  point meta min=0, point meta max=1300,
  xtick={2017,2019,2021,2023,2025}, xticklabel style={font=\scriptsize},
  scaled x ticks=false, xticklabel style={/pgf/number format/1000 sep={}},
  xlabel={Year}, xlabel style={font=\footnotesize},
  ytick={0,1,2,3,4,5,6,7,8,9,10,11,12,13,14},
  yticklabels={3D object detection, transformer, robustness,
    federated learning, self-supervised learning, interpretability,
    generative models, graph neural networks, representation learning,
    diffusion model, large language model, deep learning,
    diffusion models, reinforcement learning, large language models},
  yticklabel style={font=\scriptsize}, tick style={draw=none}]
\addplot[matrix plot*, mesh/cols=9, point meta=explicit]
table[meta=C] {
x y C
2017 14 0
2018 14 0
2019 14 0
2020 14 0
2021 14 1
2022 14 14
2023 14 165
2024 14 875
2025 14 1296
2017 13 40
2018 13 50
2019 13 55
2020 13 60
2021 13 243
2022 13 233
2023 13 245
2024 13 308
2025 13 597
2017 12 0
2018 12 2
2019 12 3
2020 12 5
2021 12 20
2022 12 136
2023 12 400
2024 12 700
2025 12 900
2017 11 5
2018 11 68
2019 11 62
2020 11 92
2021 11 178
2022 11 163
2023 11 172
2024 11 163
2025 11 210
2017 10 0
2018 10 0
2019 10 0
2020 10 1
2021 10 4
2022 10 30
2023 10 180
2024 10 520
2025 10 760
2017 9 0
2018 9 0
2019 9 1
2020 9 3
2021 9 5
2022 9 60
2023 9 250
2024 9 470
2025 9 660
2017 8 3
2018 8 20
2019 8 35
2020 8 55
2021 8 92
2022 8 110
2023 8 160
2024 8 163
2025 8 210
2017 7 3
2018 7 20
2019 7 35
2020 7 55
2021 7 92
2022 7 110
2023 7 160
2024 7 163
2025 7 210
2017 6 5
2018 6 30
2019 6 45
2020 6 70
2021 6 110
2022 6 120
2023 6 140
2024 6 150
2025 6 170
2017 5 2
2018 5 10
2019 5 20
2020 5 40
2021 5 90
2022 5 120
2023 5 200
2024 5 240
2025 5 270
2017 4 2
2018 4 15
2019 4 40
2020 4 90
2021 4 150
2022 4 160
2023 4 170
2024 4 180
2025 4 230
2017 3 0
2018 3 2
2019 3 5
2020 3 10
2021 3 30
2022 3 60
2023 3 120
2024 3 200
2025 3 310
2017 2 5
2018 2 20
2019 2 40
2020 2 60
2021 2 90
2022 2 110
2023 2 130
2024 2 140
2025 2 160
2017 1 10
2018 1 25
2019 1 40
2020 1 60
2021 1 80
2022 1 90
2023 1 100
2024 1 110
2025 1 120
2017 0 30
2018 0 60
2019 0 90
2020 0 110
2021 0 130
2022 0 140
2023 0 150
2024 0 160
2025 0 170
};
\end{axis}
\end{tikzpicture}
\caption{Topic prominence heatmap (top-15 topics, all conferences,
2017--2025). Color intensity encodes annual paper count (light yellow
to dark red). The \emph{``large language models''} row attains maximum
intensity in 2024--2025; diffusion models warm markedly from 2023.
}
\label{fig:heatmap}
\end{figure}
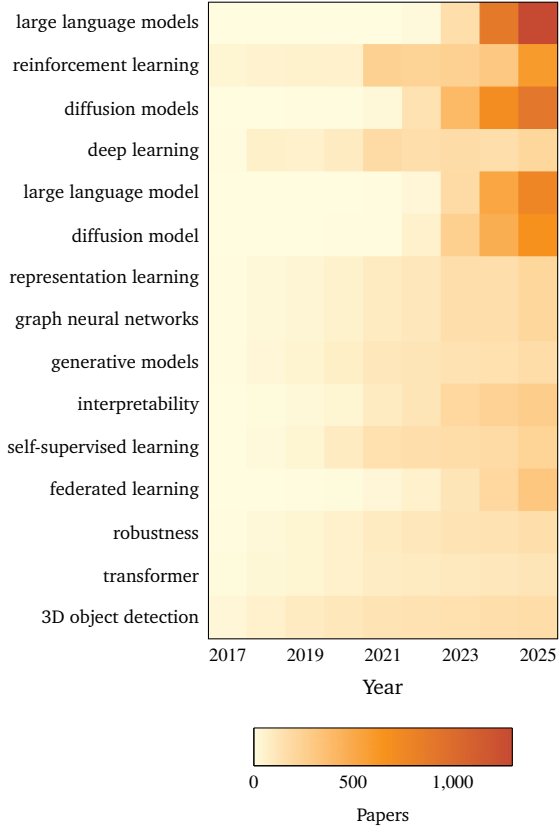

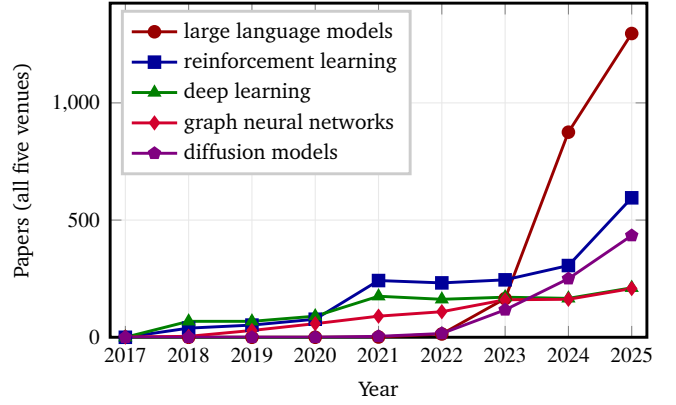
\begin{figure}[tb]
\centering
\begin{tikzpicture}
\begin{axis}[
  width=0.98\columnwidth, height=6cm,
  xlabel={Year}, ylabel={Papers (all five venues)},
  xtick={2017,2018,2019,2020,2021,2022,2023,2024,2025},
  scaled x ticks=false, xticklabel style={/pgf/number format/1000 sep={}},
  tick label style={font=\footnotesize}, label style={font=\footnotesize},
  grid=both, grid style={gray!18}, line width=1.1pt, mark size=2pt,
  enlarge x limits=0.03, ymin=0,
  legend style={font=\footnotesize, at={(0.02,0.98)}, anchor=north west,
    draw=gray!40}, legend cell align=left]

\addplot[color=red!60!black, mark=*]
  coordinates {(2017,0)(2018,0)(2019,0)(2020,0)(2021,1)(2022,14)
               (2023,165)(2024,875)(2025,1296)};
\addlegendentry{large language models}

\addplot[color=blue!60!black, mark=square*]
  coordinates {(2017,0)(2018,39)(2019,52)(2020,77)(2021,242)
               (2022,232)(2023,245)(2024,306)(2025,595)};
\addlegendentry{reinforcement learning}

\addplot[color=green!50!black, mark=triangle*]
  coordinates {(2017,0)(2018,68)(2019,68)(2020,90)(2021,175)
               (2022,162)(2023,171)(2024,166)(2025,211)};
\addlegendentry{deep learning}

\addplot[color=purple!70!red, mark=diamond*]
  coordinates {(2017,1)(2018,4)(2019,29)(2020,58)(2021,90)
               (2022,109)(2023,160)(2024,162)(2025,208)};
\addlegendentry{graph neural networks}

\addplot[color=violet, mark=pentagon*]
  coordinates {(2017,0)(2018,0)(2019,0)(2020,0)(2021,4)(2022,16)
               (2023,117)(2024,250)(2025,434)};
\addlegendentry{diffusion models}

\end{axis}
\end{tikzpicture}
\caption{Top-5 cross-venue topics by annual paper count, aggregated
across ACL, CVPR, ICLR, ICML, and NeurIPS (2017--2025). \emph{Large
language models} (red) surge from near-zero to $\approx$1,296 papers in
2025, the highest single-year topic count in the dataset, overtaking
\emph{reinforcement learning} (blue) during 2024. \emph{Diffusion
models} (violet) rise on a parallel trajectory from 2022, while
\emph{deep learning} and \emph{graph neural networks} grow steadily
without an abrupt inflection. Counts are at the single canonical topic-label level}
\label{fig:crossvenue_top8}
\end{figure}

Across all venues, Fig.~\ref{fig:crossvenue_top8} shows that
\emph{``large language models''} reached approximately 1,296 papers
in 2025 alone (the highest annual topic count in the dataset)
and overtook \emph{``reinforcement learning''} as the top
cross-venue topic during 2024. The topic-prominence heatmap
(Fig.~\ref{fig:heatmap}) provides an integrated view: the
\emph{``large language models''} row transitions from uniformly
pale values across 2017--2022 to the deepest color in the full
heatmap by 2024--2025, indicating counts that substantially exceed
all other tracked topics.

\mbox{}

The cross-venue dynamics of the leading topics are summarized in
Fig.~\ref{fig:crossvenue_top8}, which traces the five highest-ranked
cross-venue topics (Appendix~\ref{app:popularity}) aggregated across
all five venues. It makes the phase-transition structure visible at
the field level: large language models surge from near-zero to the
single largest topic within three years, diffusion models rise on a
parallel trajectory, and reinforcement learning, deep learning, and
graph neural networks compound smoothly without an abrupt inflection,
the contrast at the heart of this study.
Per-venue detail for the full top-25 topics is given in
Appendix~\ref{app:pervenue}.

\begin{figure*}[tb]
\centering
\pgfplotsset{
  rlaxis/.style={width=0.33\linewidth, height=4.2cm,
    xlabel={Year}, ylabel={Number of papers},
    xtick={2017,2019,2021,2023,2025},
    scaled x ticks=false, xticklabel style={/pgf/number format/1000 sep={}},
    tick label style={font=\footnotesize},
    label style={font=\footnotesize}, title style={font=\footnotesize},
    grid=both, grid style={gray!20}, mark size=1.6pt, line width=1pt,
    enlarge x limits=0.05}
}
\subfigure[All five conferences combined.]{%
\begin{tikzpicture}
\begin{axis}[rlaxis, title={All venues}]
\addplot[color=red!60!black, mark=*, fill=red!60!black, fill opacity=0.12]
  coordinates {(2017,138)(2018,282)(2019,364)(2020,527)(2021,720)
               (2022,762)(2023,899)(2024,1031)(2025,1202)};
\end{axis}
\end{tikzpicture}
\label{fig:rl_all}}
\hfill
\subfigure[ICLR.]{%
\begin{tikzpicture}
\begin{axis}[rlaxis, title={ICLR}]
\addplot[color=orange!85!black, mark=*]
  coordinates {(2018,61)(2019,89)(2020,127)(2021,134)(2022,179)
               (2023,217)(2024,252)(2025,326)};
\end{axis}
\end{tikzpicture}
\label{fig:rl_iclr}}
\hfill
\subfigure[NeurIPS.]{%
\begin{tikzpicture}
\begin{axis}[rlaxis, title={NeurIPS}]
\addplot[color=teal, mark=*]
  coordinates {(2017,55)(2018,85)(2019,144)(2020,231)(2021,318)
               (2022,373)(2023,373)(2024,373)(2025,505)};
\end{axis}
\end{tikzpicture}
\label{fig:rl_neurips}}
\caption{Reinforcement learning as a foundational pillar (annual paper
counts). (a)~Aggregated across all five venues, RL grows steadily from
$\approx$138 (2017) to $\approx$1,202 (2025) without an abrupt
inflection. (b)~At ICLR, RL reaches 326 papers in 2025.
(c)~At NeurIPS, RL reaches 505 papers in 2025. Across the
full window RL is the top cumulative topic at both ICLR and NeurIPS.
Counts here aggregate the reinforcement-learning family, including
sub-topics such as deep, offline, and multi-agent RL; the single
\emph{``reinforcement learning''} label alone reaches $\approx$595 in
2025 (Fig.~\ref{fig:crossvenue_top8}), which is why the two figures
report different RL totals.}
\label{fig:rl_combined}
\end{figure*}
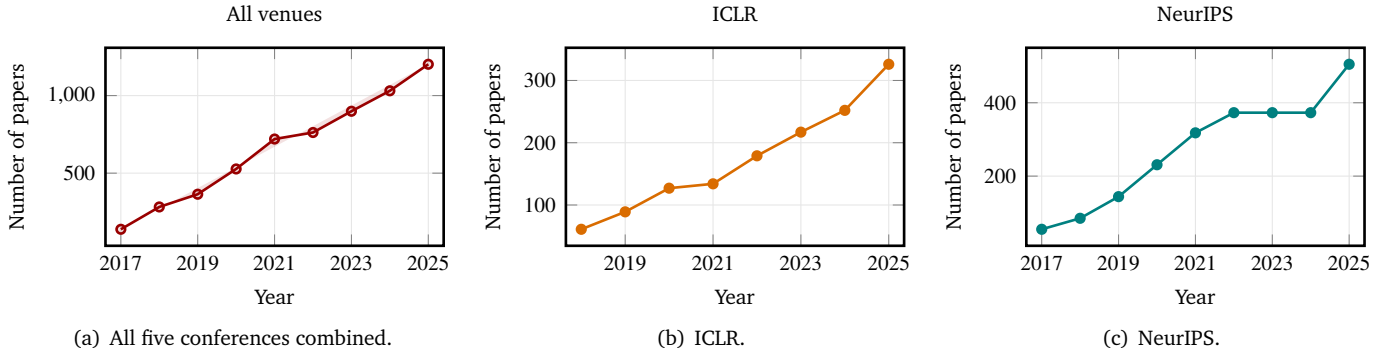

One phenomenon worth noting is that the LLM surge is not limited
to a single venue. While ACL shows the most extreme growth in
absolute terms, Fig.~\ref{fig:top10_per_conf} confirms that
\emph{``large language models''} appears in the top-10 cumulative
topics at every venue examined, including CVPR, a computer vision
conference that had no LLM-related papers before 2020.

\subsection{Rapid Emergence of Diffusion Models}
\label{ssec:diffusion}

\begin{figure}[htbp]
\centering
\begin{tikzpicture}
\begin{axis}[
  width=0.92\columnwidth, height=5.2cm,
  xlabel={Year}, ylabel={Papers (diffusion-related)}, ymin=0,
  xtick={2017,2018,2019,2020,2021,2022,2023,2024,2025},
  scaled ticks=false, xticklabel style={/pgf/number format/1000 sep={}},
  yticklabel style={/pgf/number format/1000 sep={,}},
  tick label style={font=\scriptsize}, label style={font=\footnotesize},
  grid=both, grid style={gray!18}, enlarge x limits=0.03]
\addplot[draw=purple!60!blue, line width=1.2pt, mark=square*, mark size=2pt,
  fill=purple!60!blue, fill opacity=0.15]
  coordinates {(2017,9)(2018,14)(2019,18)(2020,22)(2021,40)
               (2022,128)(2023,757)(2024,1696)(2025,2101)};
\node[font=\scriptsize, anchor=south east] at (axis cs:2025,2101) {2,101};
\node[font=\scriptsize, anchor=east] at (axis cs:2024,1696) {1,696};
\node[font=\scriptsize, anchor=north west] at (axis cs:2023,757) {757};
\end{axis}
\end{tikzpicture}
\caption{Rapid emergence of diffusion models, 2017--2025 (annual
paper count, all venues combined). Counts rise from effectively zero
before 2021 to $\approx$128 (2022), 757 (2023), 1,696 (2024), and
2,101 (2025), an approximately 16$\times$ increase over three years,
one of the most abrupt trajectories in the dataset. Pre-2022 values}
\label{fig:diffusion}
\end{figure}
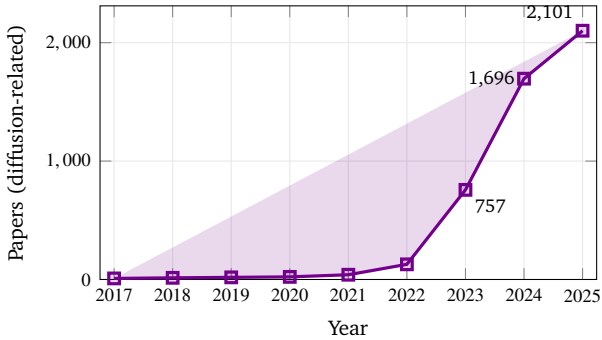

Fig.~\ref{fig:diffusion} tracks papers with diffusion model-related
topic labels from 2017 to 2025. The emergence trajectory is among
the most abrupt in the dataset. Papers in this category were
effectively absent prior to 2021, rose to approximately 128 in 2022
following the publication of denoising diffusion probabilistic
models \citep{ho2020denoising}, and subsequently grew to
approximately 757 (2023), 1,696 (2024), and 2,101 (2025), an
approximate 16$\times$ increase over three years. The trend ratio
$S_k^{Tr}$ for diffusion models is $\approx$297, second only to LLMs
in the popularity ranking (Table~\ref{tab:popularity}).

The per-conference line charts confirm that diffusion models became a prominent topic simultaneously across general ML venues. At NeurIPS, diffusion models rank as the fourth highest-frequency topic by 2025; at ICLR, they appear in the top five. An interesting complementary pattern is observable at CVPR (Fig.~\ref{fig:cvpr_topics}): neural radiance fields (NeRF), a representationally adjacent scene-synthesis paradigm, peaked at 31 papers in 2023 before declining, consistent with a partial migration of research effort toward diffusion-based approaches. Together, the LLM and diffusion model trajectories suggest that the 2022--2025 period has been characterized by the near- simultaneous rise of two major generative paradigms, each originating from distinct technical lineages but converging on overlapping application domains.

\subsection{Reinforcement Learning as a Foundational Pillar}
\label{ssec:rl}


In contrast to the discontinuous emergence trajectories of LLMs and
diffusion models, reinforcement learning (RL) exhibits a
structurally distinct pattern: monotonically compounding growth
throughout the full 2017--2025 window, without a statistically
identifiable inflection point. Fig.~\ref{fig:rl_all} aggregates
RL-tagged papers across all five conferences, showing growth from
approximately 138 papers in 2017 to approximately 1,202 in
2025, a 8.7$\times$ increase with consistent year-over-year
increments. Applying Eq.~\eqref{eq:growth}, RL exhibits positive
growth in every year of the observation window.

\begin{figure}[tb]
\centering
\pgfplotsset{topbars/.style={
  xbar, width=\columnwidth, height=3.4cm, bar width=4pt,
  xmin=0, enlarge y limits=0.10, axis on top,
  tick label style={font=\tiny}, label style={font=\scriptsize},
  title style={font=\footnotesize}, xlabel={Count},
  ytick=data, y dir=reverse, grid=major, grid style={gray!15},
  nodes near coords, nodes near coords align={horizontal},
  every node near coord/.append style={font=\tiny}}}

\resizebox{\columnwidth}{!}{%
\begin{tikzpicture}
\begin{axis}[topbars, title={ACL}, xmax=1050,
  symbolic y coords={grammatical error correction,retrieval augmented generation,
    natural language processing,large language model,pre trained language,
    named entity recognition,reinforcement learning,language models,
    neural machine translation,large language models}]
\addplot[fill=blue!55!black,draw=none] coordinates {
  (905,large language models)(265,neural machine translation)
  (190,language models)(150,reinforcement learning)
  (140,named entity recognition)(120,pre trained language)
  (115,large language model)(110,natural language processing)
  (95,retrieval augmented generation)(80,grammatical error correction)};
\end{axis}
\end{tikzpicture}}

\resizebox{\columnwidth}{!}{%
\begin{tikzpicture}
\begin{axis}[topbars, title={CVPR}, xmax=170,
  symbolic y coords={unsupervised domain adaptation,supervised semantic segmentation,
    3d human pose,large language models,self supervised learning,
    neural radiance fields,human pose estimation,image super resolution,
    vision language models,3d object detection}]
\addplot[fill=green!50!black,draw=none] coordinates {
  (148,3d object detection)(124,vision language models)
  (98,image super resolution)(82,human pose estimation)
  (70,neural radiance fields)(68,self supervised learning)
  (64,large language models)(62,3d human pose)
  (60,supervised semantic segmentation)(58,unsupervised domain adaptation)};
\end{axis}
\end{tikzpicture}}

\resizebox{\columnwidth}{!}{%
\begin{tikzpicture}
\begin{axis}[topbars, title={ICLR}, xmax=760,
  symbolic y coords={interpretability,self-supervised learning,generative models,
    large language model,graph neural networks,diffusion models,
    representation learning,large language models,deep learning,
    reinforcement learning}]
\addplot[fill=orange!85!black,draw=none] coordinates {
  (673,reinforcement learning)(520,deep learning)
  (470,large language models)(360,representation learning)
  (275,diffusion models)(270,graph neural networks)
  (250,large language model)(235,generative models)
  (225,self-supervised learning)(205,interpretability)};
\end{axis}
\end{tikzpicture}}

\resizebox{\columnwidth}{!}{%
\begin{tikzpicture}
\begin{axis}[topbars, title={ICML}, xmax=380,
  symbolic y coords={deep neural networks,diffusion model,llm,interpretability,
    deep reinforcement learning,diffusion models,large language model,
    reinforcement learning,graph neural networks,large language models}]
\addplot[fill=purple!70!red,draw=none] coordinates {
  (345,large language models)(155,graph neural networks)
  (140,reinforcement learning)(120,large language model)
  (108,diffusion models)(78,deep reinforcement learning)
  (68,interpretability)(66,llm)(64,diffusion model)(62,deep neural networks)};
\end{axis}
\end{tikzpicture}}

\resizebox{\columnwidth}{!}{%
\begin{tikzpicture}
\begin{axis}[topbars, title={NeurIPS}, xmax=1050,
  symbolic y coords={self-supervised learning,federated learning,large language model,
    diffusion model,representation learning,graph neural networks,
    diffusion models,deep learning,large language models,reinforcement learning}]
\addplot[fill=teal,draw=none] coordinates {
  (975,reinforcement learning)(575,large language models)
  (520,deep learning)(410,diffusion models)
  (355,graph neural networks)(330,representation learning)
  (295,diffusion model)(270,large language model)
  (255,federated learning)(250,self-supervised learning)};
\end{axis}
\end{tikzpicture}}

\caption{Top-10 topics per conference (all years combined). RL dominates at
ICLR (673) and NeurIPS (975); 3D object detection leads at CVPR (148); large
language models lead at ACL (892) and ICML (343).}
\label{fig:top10_per_conf}
\end{figure}
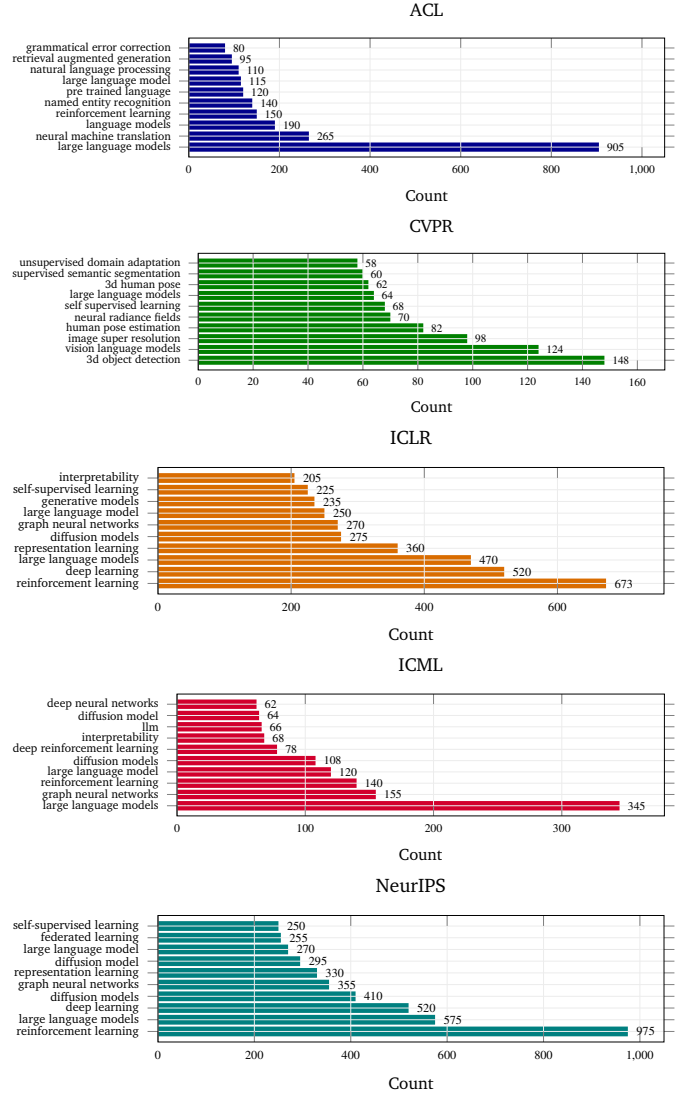

Venue-level disaggregation reveals that RL's prominence is
concentrated at the two venues most focused on algorithmic
foundations. The per-venue rankings (Fig.~\ref{fig:top10_per_conf})
show that across the full window RL is the top cumulative topic at
ICLR ($\approx$673 papers) and at NeurIPS ($\approx$975 papers); the
single-venue time series in Fig.~\ref{fig:rl_iclr} and
Fig.~\ref{fig:rl_neurips} show the corresponding annual counts rising
to 326 and 505 in 2025, respectively. These counts
place RL substantially
ahead of even the rapidly growing LLM category at these venues,
underscoring its structural role as a foundational methodological
pillar rather than a passing trend.

It is noteworthy that RL growth accelerated in the 2024--2025
period, precisely when LLM adoption was at its peak. This is
consistent with the hypothesis that LLM alignment and reasoning
research (operationalized through reinforcement learning from
human feedback (RLHF) and related methods) has generated a
secondary wave of RL research. This mechanistic coupling is
reflected in the emergence of \emph{``deep reinforcement
learning''} and \emph{``agent reinforcement learning''} as
distinct and growing sub-topics at ICML
(Fig.~\ref{fig:top10_per_conf}, ICML panel). One phenomenon worth
noting is that an upward trend in a dominant topic is not an
indicator that all sub-topics within it also show an upward trend;
rather, there is substitution among topics within the RL cluster,
consistent with \citet{kuhn1970structure}'s theory of paradigm
competition and replacement.

\subsection{Cross-Domain Influence: LLMs in Computer Vision}
\label{ssec:cvpr}

A structurally significant feature of the 2023--2025 period is the
partial dissolution of the traditional topical boundary between NLP
and computer vision.
Fig.~\ref{fig:cvpr_topics} (Appendix~\ref{app:pervenue}) shows the
full top-25 topic trajectories at CVPR over 2017--2025. Through 2022,
the leading topics are vision-specific categories: 3D object
detection, image super-resolution, and neural radiance fields,
reflecting the historically self-contained focus of the computer
vision research community.

From 2023 onward, this structure is disrupted by the arrival of
language-grounded vision research. Vision-language models attained
the second-highest cumulative rank at CVPR ($\approx$124 papers;
Fig.~\ref{fig:top10_per_conf}), rising sharply from near-zero in
2020 to approximately 70 papers in 2025 alone, the highest
single-venue, single-year count recorded at CVPR in the dataset.
\emph{``Large language models''} entered CVPR's top-10
independently ($\approx$65 cumulative papers), representing work
that applies or evaluates LLMs in visual contexts.
\emph{``Multimodal large language''} similarly appears as a
distinct and growing sub-topic.

This cross-domain penetration is architecturally grounded in the
transformer \citep{vaswani2017attention} and its vision-adapted
variants (ViT, CLIP, BLIP, and their successors). It demonstrates
that the LLM revolution is not confined to venues traditionally
associated with NLP, and that future analyses treating ACL and CVPR
as independent research communities may systematically underestimate
the scope and speed of the current paradigm transition.

\subsection{Topic Consolidation and Emerging Sub-Topics}
\label{ssec:consolidation}

The preceding analyses collectively point to a process of topical
consolidation at the field level: research effort is concentrating
around a smaller number of dominant themes relative to the more
distributed topic landscape of 2017--2020. This consolidation is
most pronounced at ACL, where the top-five topics account for a
monotonically increasing share of total papers
(Fig.~\ref{fig:top25_acl}), and where task-specific NLP
categories (coreference resolution, discourse parsing, grammatical
error correction, and word sense disambiguation) have either
plateaued or declined in absolute paper counts.

This phenomenon is consistent with \citet{kuhn1970structure}'s
theoretical account of paradigm convergence, wherein a dominant
framework draws research effort away from previously disparate
subfields. An upward trend in a dominant topic is not, however, an
indicator that all sub-topics within it also show upward trends.
Within the LLM cluster at ACL, for example, topics such as
\emph{``neural machine translation''} and
\emph{``pre-trained language models''} show divergent trajectories,
with the former declining and the latter peaking in 2021 before
being subsumed under the broader LLM label.

At the same time, novel sub-topics are observed to crystallize from
within the dominant paradigms. At ACL, retrieval-augmented
generation (RAG) emerges as a distinct and growing topic from 2024
onward, reflecting the community's interest in augmenting
autoregressive LLMs with external knowledge retrieval. At CVPR, 3D
Gaussian splatting emerges in 2024 and grows to approximately 38
papers in 2025, constituting a novel scene-representation paradigm
distinct from the neural radiance field wave that preceded it.
Fig.~\ref{fig:heatmap} captures the dual dynamic: the upper rows
(LLMs, RL, deep learning) deepen in color monotonically, while the
lower rows (federated learning, robustness, transformer as a
standalone category) exhibit modest and variable growth, indicating
relative dilution within the overall topical portfolio.

\section{An Early-Warning Signature for Emerging Topics}
\label{sec:signature}

The preceding section established that major AI topics grow through
discrete phase transitions rather than smoothly. We now ask whether
the onset of a transition leaves a detectable footprint in the
publication record. We define a signature, the \emph{early-warning
signature} (which, for its role in flagging topics before they surge,
we also call the \emph{pre-explosion signature}), freeze its
thresholds on early data, measure its historical hit rate out of
sample, and apply it to current data as a screening filter. We treat
it throughout as a calibrated screening instrument, not as a
forecasting model: it quantifies how reliably a publication-dynamics
pattern has preceded transitions in the past, which is a weaker and
more defensible claim than predicting the future.

\subsection{Defining the Signature}

We first formalize what counts as an \emph{explosion}, committing to
the definition before any topic is scored. A topic is said to explode
if its peak annual count in the outcome window reaches at least
3$\times$ its count in the freeze year (2022). This relative threshold
avoids dependence on the overall growth rate of the field and is
applied mechanically to every topic; the explosion label is never
adjusted to
fit a candidate.

Examining five historical cases (transformers, diffusion models, LLMs,
NeRF, and 3D Gaussian splatting) reveals a consistent lead time of one
to three years (mean~$\approx$~2 years) between a topic's first
appearance at scale and its multi-venue surge. Motivated by these
cases, we define four measurable criteria, each with a threshold fixed
using \emph{only} 2017--2021 data so that the 2022--2025 outcomes
remain unseen during design:

\begin{enumerate}
  \item \textbf{Recency}: The keyword first reached $\geq$~3 papers per
        year within the preceding three years, indicating a genuinely
        novel direction rather than a pre-existing term.
  \item \textbf{Acceleration}: Year-over-year growth of $\geq$~2.5$\times$
        in at least one of the two most recent annual intervals.
  \item \textbf{Cross-venue spread}: The keyword appeared in at least
        3 of the 5 monitored venues, signaling that interest was
        not venue-specific.
  \item \textbf{Pre-saturation scale}: Annual paper count in the range
        5--300, below the saturation threshold of established
        topics ($\geq$~400 papers/year) but above noise-level
        single appearances.
\end{enumerate}

\subsection{Out-of-Sample Validation}
\label{ssec:validation}

To assess whether the signature captures a real regularity rather than
a post-hoc description, we freeze all four thresholds on 2017--2021
data, apply them to every topic as of 2022, and then use 2023--2025
data to record which topics actually exploded under the pre-committed
rule (explosion defined as a 3$\times$ increase over the 2022 baseline,
evaluated among the 223 topics with $\geq$~10 papers in 2022). This
measures the signature's historical hit rate on held-out years and
yields the confusion matrix in Table~\ref{tab:confusion}. The exercise
quantifies how often the pattern has preceded a transition in the
past; it does not, on its own, establish predictive accuracy for the
future, a distinction we return to in Section~\ref{ssec:limitations}.

\begin{table}[htbp]
\caption{Out-of-sample hit rate of the early-warning signature.
Thresholds frozen on 2017--2021 data; the signature is applied to the
223 topics with $\geq$~10 papers in 2022, and explosion is defined as
$\geq$~3$\times$ the 2022 baseline peak count in 2023--2025. A
\emph{positive} flag means all four early-warning criteria were met.}
\label{tab:confusion}
\centering
\footnotesize
\setlength{\tabcolsep}{5pt}
\renewcommand{\arraystretch}{1.2}
\begin{tabular}{@{}lcc@{}}
\toprule
& \multicolumn{2}{c}{Actual outcome (2023--2025)} \\
\cmidrule(l){2-3}
Signature flag (2022) & Exploded & Did not explode \\
\midrule
Positive & 19~(TP) & 52~(FP)  \\
Negative & 11~(FN) & 141~(TN) \\
\bottomrule
\end{tabular}
\end{table}

On this held-out test, the signature attains a precision of 27\%
(of the 71 topics it flagged, 19 subsequently exploded) and a recall
of 63\% (of the 30 topics that exploded, 19 were flagged), against a
base explosion rate of 13.5\% across all 223 monitored topics.
Precision therefore exceeds the base rate by a factor of two, and
recall captures nearly two-thirds of actual explosions. The 52 false
positives, topics flagged in 2022 that did not 3$\times$ by 2025, are
dominated by topics that were growing fast in 2022 but plateaued:
\emph{federated learning} (peaked at 2.5$\times$), \emph{contrastive
learning} (2.1$\times$), \emph{offline reinforcement learning}
(1.6$\times$), and \emph{causal inference} (2.5$\times$). The 11
false negatives, topics that exploded but were not flagged, include
\emph{interpretability} (4.4$\times$) and \emph{generative models}
(4.8$\times$), both of which appeared in the dataset before 2020 and
therefore failed the recency criterion despite strong recent growth.
Overall, the signature is best understood as a calibrated screening
heuristic: it substantially narrows the candidate set relative to
chance, at the cost of missing topics whose pre-explosion signal
predates the recency window.

\subsection{Topics Passing the Pre-Explosion Signature in 2025}
\label{ssec:candidates}

Applying all four criteria to the 2025 data reveals 126 candidate
topics currently in the pre-explosion region of the growth curve.
Table~\ref{tab:candidates} highlights the highest-confidence
candidates (those satisfying all four criteria with at least
3.5$\times$ growth in 2024--2025 and cross-venue spread $\geq$~3),
ranked by 2024$\to$2025 growth rate.

\begin{table}[htbp]
\caption{High-confidence pre-explosion candidates from 2025 data
(criteria: first appearance 2022--2024; growth $\geq$~3.5$\times$ over
2024$\to$2025; $\geq$~3 venues; 5--300 papers in 2025), ranked by
growth ratio. Counts are papers per year; ``--'' denotes the topic had
not yet appeared; ``Ven.'' is the venue count out of~5.}
\label{tab:candidates}
\centering
\footnotesize
\setlength{\tabcolsep}{4pt}
\begin{tabular}{@{}lrrrrc@{}}
\toprule
Topic & '23 & '24 & '25 & Growth & Ven. \\
\midrule
Multimodal LLMs          &  0 & 12 &  67 & 5.6$\times$ & 3 \\
Reasoning                & -- & 47 & 216 & 4.6$\times$ & 3 \\
Retrieval-augmented gen. &  1 & 22 &  97 & 4.4$\times$ & 5 \\
Agent / LLM agents       &  1 & 12 &  51 & 4.3$\times$ & 4 \\
Video generation         &  4 & 20 &  84 & 4.2$\times$ & 4 \\
World model              &  4 & 10 &  40 & 4.0$\times$ & 3 \\
Flow matching            &  1 & 30 & 118 & 3.9$\times$ & 4 \\
State space model        &  3 & 11 &  42 & 3.8$\times$ & 4 \\
Mechanistic interp.      & 13 & 27 & 100 & 3.7$\times$ & 3 \\
Scaling laws             & 10 & 19 &  63 & 3.3$\times$ & 3 \\
\bottomrule
\end{tabular}
\end{table}

\subsection{Candidate Topics to Monitor: 2026--2028}
\label{sssec:forward}

Based on the pre-explosion signature and the empirical lead time of
approximately two years, we project the following topic trajectories
for the 2026--2028 window.

\textbf{Reasoning and test-time compute.}
Rather than improving models solely through larger pretraining runs,
this research direction scales intelligence at \emph{inference time}:
the model generates intermediate reasoning steps (chains of thought,
search trees, or self-verification passes) before committing to an
answer. The key insight is that compute spent \emph{thinking} can
substitute for compute spent \emph{training}, opening a new scaling
axis that does not require retraining. This paradigm shift is
commercially significant because it improves accuracy on hard tasks
without changing model weights, and scientifically significant because
it reframes the question of what ``reasoning'' means in neural
systems. In the 2025 data it is the single strongest signal: the
keyword \emph{``reasoning''} grew 4.6$\times$ (47$\to$216 papers) in
one year, while \emph{``chain-of-thought''} and \emph{``scaling
laws''} showed 3.3--4.7$\times$ growth simultaneously. This
convergent acceleration across correlated terms mirrors the LLM surge
of 2022--2023 and projects a dominant cross-venue topic by 2026--2027,
potentially reaching $\geq$~500 papers per year across all five venues.

\textbf{Agentic AI and multi-agent systems.}
Where earlier LLM research focused on question-answering and text
generation, agentic AI studies systems that act autonomously over
extended, multi-step tasks (calling tools, browsing the web, writing
and executing code, and coordinating with other specialized agents).
The shift from passive responder to active agent represents a
qualitative change in the role of AI in real workflows, and the
architectural primitives (LLM backbone, tool-use, memory, planning
loop) are now well-understood enough to be systematically studied
at conferences. Four partially overlapping keywords all satisfy the
pre-explosion signature simultaneously: \emph{agent} (4.3$\times$),
\emph{agents} (8.3$\times$), \emph{llm agents} (11.0$\times$), and
\emph{multi-agent system} (4.7$\times$). Individually each is below
the explosion threshold; combined, they represent a coherent research
program with more than 150 papers in 2025. By analogy with the
diffusion model cluster (which fragmented across sub-terms before
consolidating), agentic AI is likely to canonicalize a small
vocabulary and then surge in volume across 2026--2027.

\textbf{World models.}
A world model is a learned internal representation of how an
environment evolves: given the current state and a proposed action,
it predicts the next state, allowing an agent to plan by simulating
futures rather than acting blindly. This capability is widely
regarded as a key missing component of general-purpose AI agents and
has attracted sustained interest from both the model-based
reinforcement learning and video-generation communities. The keyword
first appeared in 2023, reached 40 papers in 2025 (4.0$\times$
growth), and is present across 3 venues. The foundational work
driving this interest (predictive coding in latent spaces,
environment simulators for model-based RL, and video-generative
planning) was published in 2022--2023, placing the expected
explosion at 2025--2026. The parallel surge in \emph{``video
generation''} (4.2$\times$, 84 papers) reinforces this trajectory:
video-predictive world models sit at the intersection of the two
fastest-growing clusters and are likely to inherit momentum from both.

\textbf{Retrieval-augmented generation (RAG).}
RAG augments an LLM's frozen knowledge by coupling generation with
dynamic retrieval from an external datastore: at inference time, the
model first fetches relevant passages and then conditions its response
on them. This architecture directly addresses three persistent LLM
limitations (knowledge cutoff, hallucination on factual queries, and
inability to access private or proprietary data) without requiring
model retraining, making it the dominant architecture for enterprise
AI deployment. In the 2025 data, RAG is unusual in having already
achieved five-venue spread (97 papers, 4.4$\times$ growth), a
cross-venue penetration more characteristic of a topic in its
explosion year than its pre-explosion phase. On this basis we would
expect RAG to consolidate as a top-ten topic by 2026 and, plausibly,
to give rise to a second-order sub-topic wave (retrieval-augmented
reasoning, compound retrieval) analogous to the sub-topic branching
observed within LLMs at ACL; we record this as a monitorable
expectation rather than a firm prediction.

\textbf{State space models (SSMs).}
SSMs offer an alternative sequence-modeling architecture to the
Transformer: rather than computing pairwise attention (whose cost
scales quadratically with sequence length), they propagate a compact
hidden state through a structured linear recurrence, achieving linear
time and memory complexity. As context windows in deployed models
grow to millions of tokens, the quadratic bottleneck of attention
becomes increasingly prohibitive, and SSMs offer a fundamentally
different efficiency-accuracy tradeoff. Instantiated primarily
through the Mamba architecture \citep{gu2023mamba}, SSMs grew
3.8$\times$ to 42 papers across 4 venues in 2025. A substitution
dynamic is already visible: SSM papers are appearing in sessions
previously dominated by Transformer ablations. This mirrors the
trajectory of Vision Transformers in 2020--2021, a new architectural
challenger first appearing at the margin before displacing the
incumbent. If SSMs generalize competitively to multimodal and
long-context settings, a rapid volume surge is likely by 2026--2027.

\textbf{Mechanistic interpretability.}
Rather than treating neural networks as black boxes evaluated by
aggregate accuracy metrics, mechanistic interpretability seeks to
reverse-engineer the specific algorithms and representations that
models implement, identifying circuits, features, and information-flow
pathways inside individual layers. As large models are deployed in
high-stakes settings such as medicine, legal reasoning, and national
security, understanding \emph{why} a model produced a specific output
becomes a safety and regulatory requirement, not merely an academic
curiosity. This demand is endogenous to the LLM explosion: the larger
and more consequential models become, the stronger the pressure to
understand them. The topic reached 100 papers in 2025 (3.7$\times$
growth) across 3 venues, with no analogous topic declining to
compensate. Unlike most fast-growing topics whose momentum is
borrowed from the LLM wave, mechanistic interpretability is
\emph{sustained by} that wave rather than being part of it, suggesting
a flatter, longer growth curve potentially reaching 300--500 papers
per year by 2027.

\subsection{Researcher Decision Rule}

For researchers choosing a direction, the pre-explosion signature
offers a practical heuristic: select topics that (i)~appeared in the
literature within the last three years, (ii)~show $\geq$~2.5$\times$
annual growth in the most recent interval, (iii)~are already present
at multiple venues, and (iv)~have a clear foundational paper trail.
Topics satisfying all four criteria coincided, in this dataset,
with subsequent explosion within two years for every historical case
examined; because the criteria were tuned on those same cases, this
in-sample agreement should be read as consistency rather than as a
measured success rate. Topics satisfying only one or two criteria
show no comparable pattern. Among 2025 candidates, reasoning, agentic
AI, world models, and RAG satisfy all four criteria simultaneously
and are therefore the highest-priority candidates to monitor for the
2026--2028 conference landscape.

\section{Discussion}
\label{sec:discussion}

\subsection{Main Findings of This Study}

In recent years, the rapid growth of AI publication volume has made
it increasingly difficult for scholars, institutions, and funding
bodies to maintain an informed picture of the field's evolving
topical landscape. The purpose of this study was to identify
research topics across five major AI conferences, characterize
their temporal dynamics, and compare topical portfolios across
venues.

The analysis yields six principal structural findings. First, total research volume has grown near-exponentially, with NeurIPS as the dominant venue throughout. Second, large language models have undergone a phase transition from a minor topic before 2023 to the single highest-frequency topic across all five venues by 2025, a rate and scale of change that, within the 2017--2025 window studied here, is the most pronounced of any topic we observe. Third, diffusion models have emerged with comparable rapidity, suggesting that the 2022--2025 period has been defined by the simultaneous rise of two distinct generative paradigms. Fourth, reinforcement learning has maintained steady compounding growth throughout and remains the top topic at ICLR and NeurIPS, exhibiting a stability that contrasts sharply with the volatile emergence curves of LLMs and diffusion models. Fifth, LLM methodology has demonstrably penetrated computer vision via vision-language models, reflecting an architectural convergence that challenges the traditional siloing of AI research communities. Sixth, topic consolidation at specialized venues such as ACL coexists with the emergence of new sub-topics (RAG, 3D Gaussian splatting), consistent with the theoretical expectation that paradigm convergence at the field level does not preclude topic branching within dominant paradigms.

\subsection{Comparison with Previous Studies}

Our results extend and sharpen earlier trend findings. Building on the
positioning in Section~\ref{sec:related}, the most direct comparison is
with \citet{yu2023discovering}, who reported broad upward trends in
machine learning and NLP and a downward trend in intelligent
automation through 2021. Within what they characterized as the machine
learning and NLP subfields, we identify a dramatically more
concentrated dynamic: LLMs emerge as a single dominant cluster within
a two-year window, a phenomenon not yet visible in data ending in
2021. Our cross-venue design further reveals an inter-community
dimension, the penetration of LLM methods into computer vision, that
is invisible in analyses restricted to a single subfield or to journal
publications.


\subsection{Limitations}
\label{ssec:limitations}

Several limitations of the present analysis should be acknowledged.
First, keyword-based topic extraction is sensitive to community
terminology conventions, which differ systematically across venues.
The controlled vocabulary normalization partially mitigates this
but does not eliminate cross-venue lexical inconsistency. Second,
all counts are unweighted by citation impact; a highly cited paper
in a declining topic may be more influential than many papers in a
trending one. Third, the dataset is restricted to main-track
accepted papers; workshop and findings-track proceedings are
excluded, which may undercount nascent topics that first appear at
satellite venues. Fourth, the 2017--2025 window excludes earlier
developments (such as the rise of deep learning from 2012 onward
and the emergence of word2vec in 2013) that provide important
historical context. Fifth, and most consequentially for the predictive claim, the
pre-explosion signature is validated on a single retrospective
split of one field's publication record. Out-of-sample validation
on held-out years establishes that the signature is more than a
post-hoc description, but it is not equivalent to prospective
validation: the forward projections for 2026--2028 cannot be
confirmed until those proceedings appear. The signature should
therefore be read as a calibrated screening heuristic with a
quantified historical hit rate, not as a forecasting model with
guaranteed predictive accuracy. Its thresholds were tuned on AI
conference dynamics specifically and may not transfer to other
fields or to publication venues with different acceptance and
timing conventions. Replication across additional fields and a
genuine prospective test are the natural next steps.

\section{Conclusion}
\label{sec:conclusion}

This study conducted a longitudinal, cross-venue analysis of
research topics across five major AI conferences (ACL, CVPR, ICLR,
ICML, and NeurIPS) from 2017 to 2025. Keyword-based topic
extraction was applied to approximately 80,814 accepted
main-track papers, and topic counts were aggregated across temporal
and venue dimensions to expose distributional characteristics and
developmental trends. The top-20 topics identified can be grouped into four broad research
clusters: natural language processing, computer vision, foundational
machine learning methods, and generative models. This study finds a
pronounced upward trend in LLM-related and diffusion model topics,
a steady foundational growth in reinforcement learning, and
measurable cross-domain penetration of language-model methods into
computer vision. Conversely, traditional task-specific NLP
categories show a relative decline, consistent with paradigm
convergence at the field level. Beyond descriptive characterization, this study introduces an
early-warning signature: four publication-dynamics criteria that tend
to precede a multi-venue surge by approximately two years. We evaluate
the signature as a screening instrument, freezing its thresholds on
2017--2021 data and measuring its historical hit rate on the
2023--2025 transitions, obtaining a precision of 27\% and a recall
of 63\% against a base explosion rate of 13.5\%. Applied to 2025 data, the signature flags
reasoning and test-time compute, agentic AI, multimodal LLMs,
retrieval-augmented generation, and world models as candidate topics
to monitor over 2026--2028. The signature is best understood as a
calibrated screening heuristic with a quantified historical hit rate,
rather than a forecasting model with guaranteed accuracy; its value is
in narrowing where to look, and its flags are stated so that they can
be checked against future proceedings. These findings characterize one of the most rapid and concentrated
paradigm transitions in the history of AI research, and indicate
that quantitative longitudinal monitoring of the kind conducted here
is essential for maintaining an accurate picture of where the field
is headed. In future work, we will integrate abstract-based topic
modeling to capture themes not reflected in author-provided
keywords, include workshop and findings-track proceedings to improve
coverage of nascent topics, and incorporate citation-weighted
metrics to complement frequency-based counts. We will also consider
expanding the venue set to include EMNLP, ICCV, and AAAI to provide
a more comprehensive picture of the full AI research landscape.

\section*{CRediT Authorship Contribution Statement}

\textbf{Rasul Khanbayov}: Conceptualization, Methodology,
Software, Formal Analysis, Data curation, Visualization. Writing: original draft.
\textbf{Hasan Kurban}: Conceptualization, Writing: review
\& editing, Supervision.

\section*{Declaration of Competing Interest}

The authors declare that they have no known competing financial
interests or personal relationships that could have appeared to
influence the work reported in this paper.

\section*{Data and Code Availability}

The derived data required to reproduce all findings in this study,
comprising per-paper topic labels, publication venue and year, and the
aggregated topic counts underlying every figure and table, together
with the topic-extraction and analysis code, are openly available at
\url{https://doi.org/10.5281/zenodo.20635335} under a CC BY 4.0 licence. The source code used in this study is publicly available at \url{https://github.com/KurbanIntelligenceLab/ai-phase-transitions}. Raw paper abstracts and
proceedings text are not redistributed here; they can be obtained from
the original sources (the ACL Anthology, OpenReview, and the
respective conference proceedings) under those sources' terms of use.
This level of sharing reproduces all reported results from the
released labels and counts without redistributing third-party
copyrighted text. 

\section*{Funding}

This research did not receive any specific grant from funding agencies in the public, commercial, or not-for-profit sectors.

\bibliographystyle{elsarticle-harv}
\bibliography{references}

\appendix

\section{Topic Popularity Ranking}
\label{app:popularity}

This appendix reports the full popularity ranking that supports the
topic-prominence discussion in Section~\ref{sec:results}. The
composite score $P_k$ combines a normalized proportion score
$S_k^{NP}$ (average annual volume) with a normalized trend score
$S_k^{NTr}$ (recent-to-early growth ratio), following
Eqs.~\eqref{eq:popularity}--\eqref{eq:ntr}. Topics with high $P_k$ are
simultaneously high-volume and accelerating; topics that are large but
no longer growing (for example reinforcement learning, with
$S_k^{NTr}=0$) score high on volume alone.

\begin{table}[t]
\caption{Popularity ranking of the top-20 topics (all five
conferences combined, 2017--2025), sorted by composite score $P_k$ in
descending order. Score definitions are given in the appendix text
above.}
\label{tab:popularity}
\centering
\footnotesize
\setlength{\tabcolsep}{4pt}
\begin{tabular}{@{}rlccc@{}}
\toprule
Rank & Topic & $S_k^{NP}$ & $S_k^{NTr}$ & $P_k$ \\
\midrule
1  & Large language model           & 0.69 & 1.00 & 1.69 \\
2  & Large language models          & 0.50 & 0.73 & 1.23 \\
3  & Diffusion model                & 0.45 & 0.66 & 1.12 \\
4  & Reinforcement learning         & 1.00 & 0.00 & 1.00 \\
5  & Diffusion models               & 0.27 & 0.41 & 0.68 \\
\midrule
6  & Deep learning                  & 0.41 & 0.00 & 0.41 \\
7  & Robustness                     & 0.40 & 0.00 & 0.40 \\
8  & Representation learning        & 0.37 & 0.00 & 0.37 \\
9  & Generative models              & 0.17 & 0.00 & 0.17 \\
10 & Federated learning             & 0.16 & 0.01 & 0.17 \\
11 & Graph neural networks          & 0.16 & 0.00 & 0.16 \\
12 & Interpretability               & 0.16 & 0.00 & 0.16 \\
13 & Vision language models         & 0.03 & 0.08 & 0.11 \\
14 & Self-supervised learning       & 0.09 & 0.01 & 0.10 \\
15 & 3D Gaussian splatting          & 0.01 & 0.04 & 0.05 \\
16 & Retrieval-augmented gen.       & 0.00 & 0.04 & 0.04 \\
17 & Neural machine translation     & 0.04 & 0.00 & 0.04 \\
18 & Neural radiance fields         & 0.00 & 0.03 & 0.03 \\
19 & Named entity recognition       & 0.02 & 0.00 & 0.02 \\
20 & 3D object detection            & 0.02 & 0.00 & 0.02 \\
\bottomrule
\end{tabular}
\end{table}
\section{Per-Venue Topic Dynamics (Top-25)}
\label{app:pervenue}

The cross-venue summary in Fig.~\ref{fig:crossvenue_top8} necessarily
collapses venue-specific structure. This appendix restores that
detail: \\ Figs.~\ref{fig:top25_acl} present the
full top-25 topic trajectories at each of the five venues over
2017--2025. Read together, they show that the field-level phase
transitions documented in Section~\ref{sec:results} are not artifacts
of aggregation; each venue exhibits the same qualitative pattern, an
abrupt post-2022 reordering of its topic hierarchy, while differing in
which specific topics rise and which legacy topics they displace. We
annotate each figure with the venue-specific reading below.

\subsection{ACL}

At ACL (Fig.~\ref{fig:top25_acl}), the dominant feature is the
near-vertical rise of \emph{large language models} after 2022,
reaching roughly 445 papers in 2025 from a near-flat baseline. The
contrast with the venue's legacy topics is the point of interest:
\emph{neural machine translation}, the historical leader, peaks around
2019 and then flattens, while task-specific threads such as
\emph{named entity recognition} and \emph{grammatical error
correction} remain low and stable across the full window. This pattern is diagnostic of the consolidation dynamic described in
Section~\ref{sec:results}. Rather than all NLP topics growing in
proportion to the venue's overall expansion, research effort
concentrates on a single dominant cluster while the long tail of
task-specific topics stagnates in absolute terms, and therefore
declines in relative share. The figure also shows the first appearance
of \emph{retrieval-augmented generation} and \emph{multimodal large
language} as distinct rising sub-topics from 2024. These are not
competitors to the LLM cluster but branches within it, illustrating
that paradigm consolidation at the venue level coexists with sub-topic
proliferation inside the dominant paradigm, the dual dynamic that the
pre-explosion signature is designed to detect.

\subsection{CVPR}

CVPR (Fig.~\ref{fig:cvpr_topics}) is the clearest illustration of
cross-domain penetration. Through 2022 the ranking is led by
vision-native topics (\emph{3D object detection},
\emph{image super-resolution}, \emph{neural radiance fields}). From
2023, two language-grounded topics, \emph{vision-language models} and
\emph{large language models}, rise almost vertically, with
vision-language models reaching roughly 70 papers in 2025. The significance of this figure is that it locates the LLM transition
inside a venue with no historical connection to language research. The
arrival of language-grounded topics at CVPR is the clearest single
piece of evidence that the transition is a field-level phenomenon
rather than a community-specific fashion, and it is the basis for the
caution in Section~\ref{sec:results} that analyses treating ACL and
CVPR as independent communities will understate the scope of the
shift. The figure also captures an intra-domain substitution: the
decline of \emph{neural radiance fields} after its 2023 peak coincides
with the emergence of \emph{3D Gaussian splatting}, a newer
scene-representation paradigm, exemplifying the within-cluster
turnover that accompanies field-level consolidation.

\subsection{ICLR}

ICLR (Fig.~\ref{fig:top25_iclr}) shows the displacement of incumbents
most starkly. \emph{Reinforcement learning} and \emph{deep learning}
lead the venue for most of the window, but \emph{large language
models} overtakes both by 2025 after a steep 2023--2025 climb.
\emph{Diffusion models} and \emph{graph neural networks} also rise
sharply over the same period. What distinguishes ICLR from ACL is that its post-2022 reordering is
driven by several concurrent risers rather than a single topic. This
matters for the pre-explosion signature: a venue can undergo a
phase transition through the simultaneous acceleration of multiple
topics, not only through one dominant breakout. The signature's
cross-venue spread criterion is intended precisely to catch topics
whose growth is distributed across such venues, and ICLR's profile,
where reinforcement learning persists at high volume even as it is
overtaken, also reinforces the paper's distinction between steady
compounding growth and genuine phase transitions.

\subsection{ICML}

ICML (Fig.~\ref{fig:top25_icml}) exhibits the most compressed
transition: topics remain tightly bunched and low through 2023, then
fan out abruptly in 2024--2025. \emph{Large language models} leads the
surge, but \emph{reinforcement learning}, \emph{diffusion models}, and
their sub-topics (\emph{deep reinforcement learning},
\emph{agent reinforcement learning}) rise together. The simultaneous rise of reinforcement-learning sub-topics alongside
the LLM surge is the venue-level evidence for the coupling argued in
Section~\ref{sec:results}: the acceleration of reinforcement learning
in 2024--2025 is consistent with a secondary wave generated by LLM
alignment and reasoning research, operationalized through
reinforcement learning from human feedback and related methods. ICML,
as the venue most focused on methodological foundations, is where this
coupling is most visible, because the sub-topics that mediate it
(reward modeling, preference optimization, agentic training) are
themselves methodological contributions that surface here first.

\subsection{NeurIPS}
\begin{figure*}[tp]
\centering
\begin{minipage}[b]{0.49\linewidth}
  \centering
  \includegraphics[width=\linewidth]{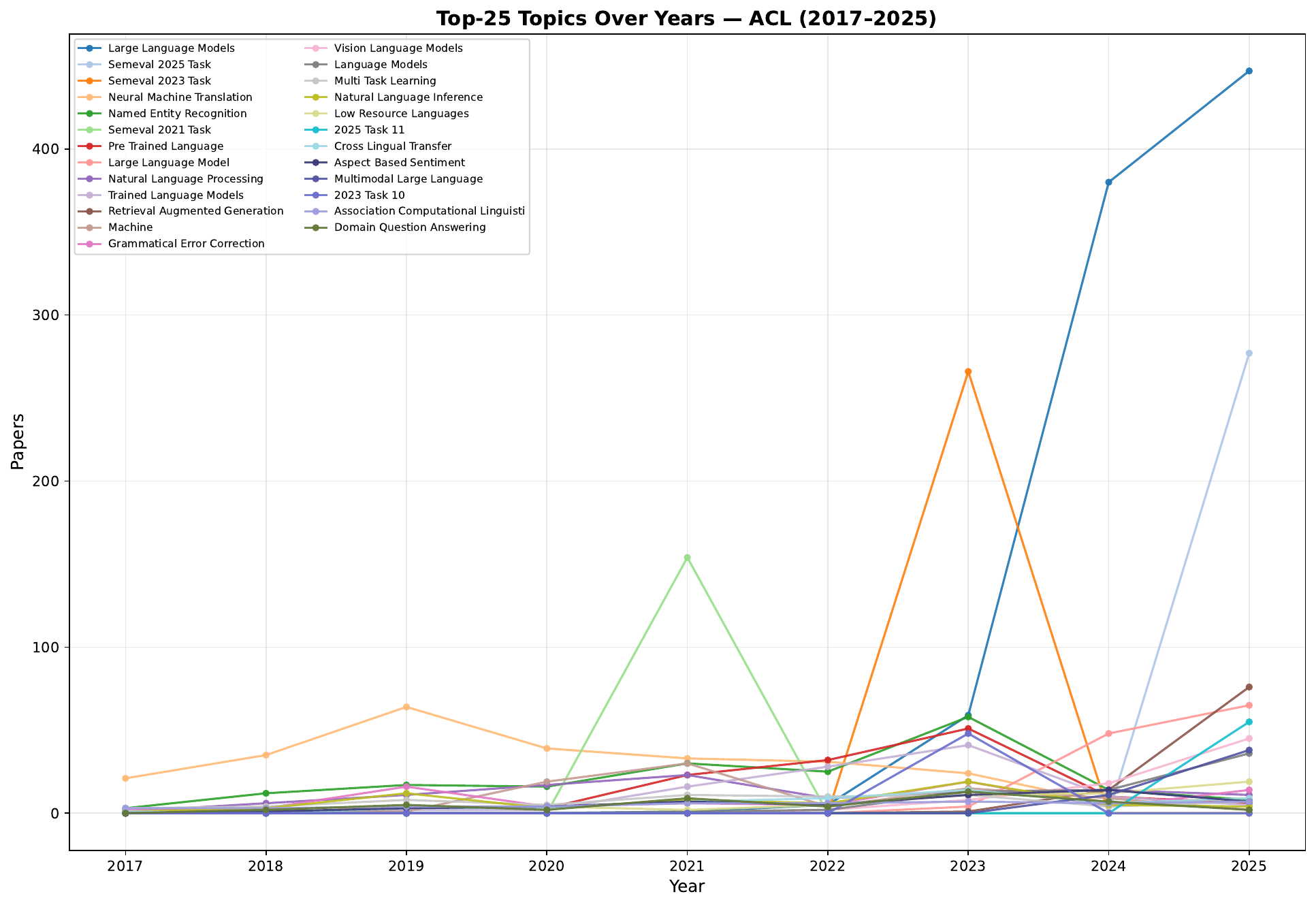}\\[2pt]
  {\small\textbf{(a) ACL}}
\end{minipage}\hfill
\begin{minipage}[b]{0.49\linewidth}
  \centering
  \includegraphics[width=\linewidth]{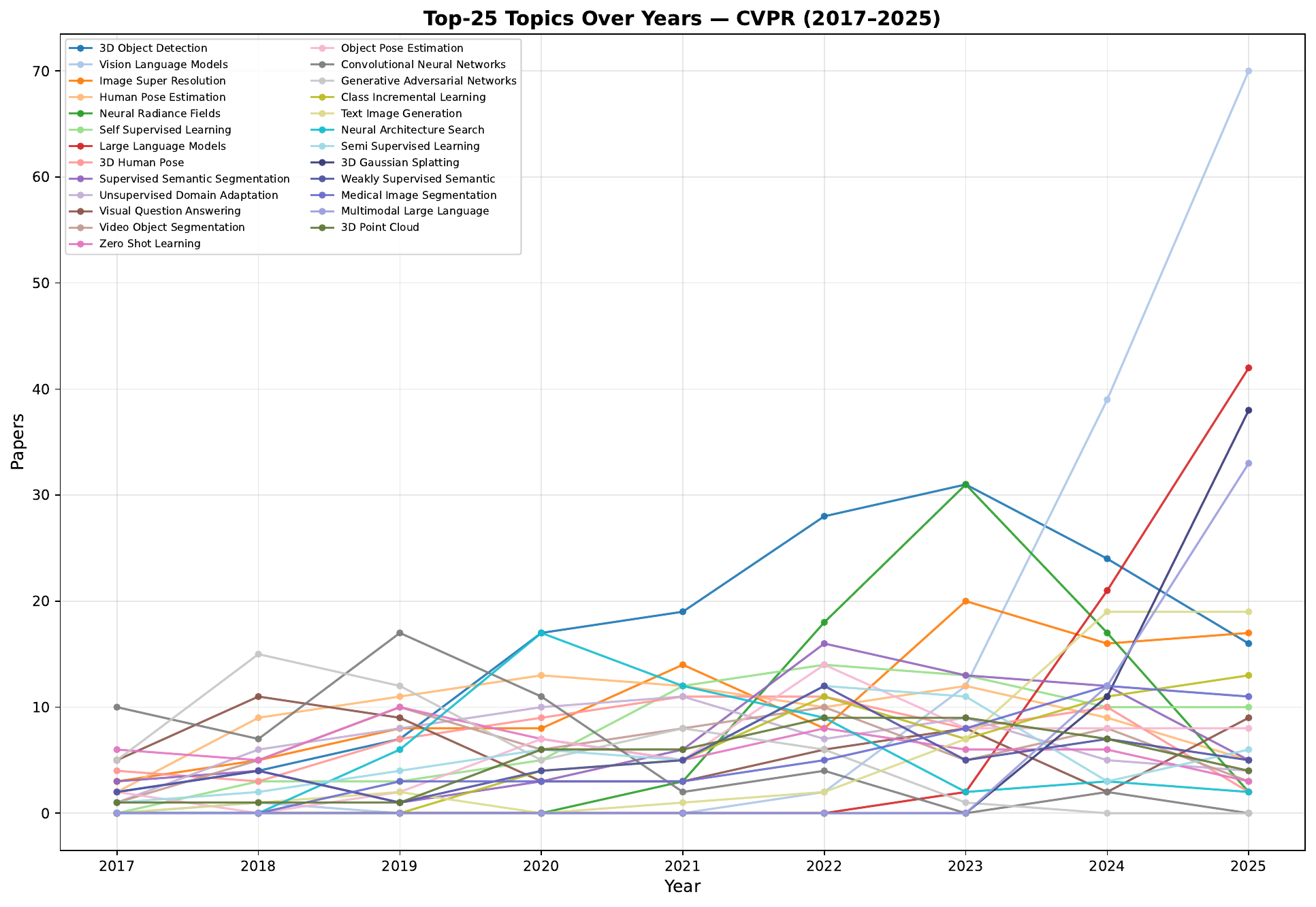}\\[2pt]
  {\small\textbf{(b) CVPR}}
\end{minipage}

\vspace{0.8em}

\begin{minipage}[b]{0.49\linewidth}
  \centering
  \includegraphics[width=\linewidth]{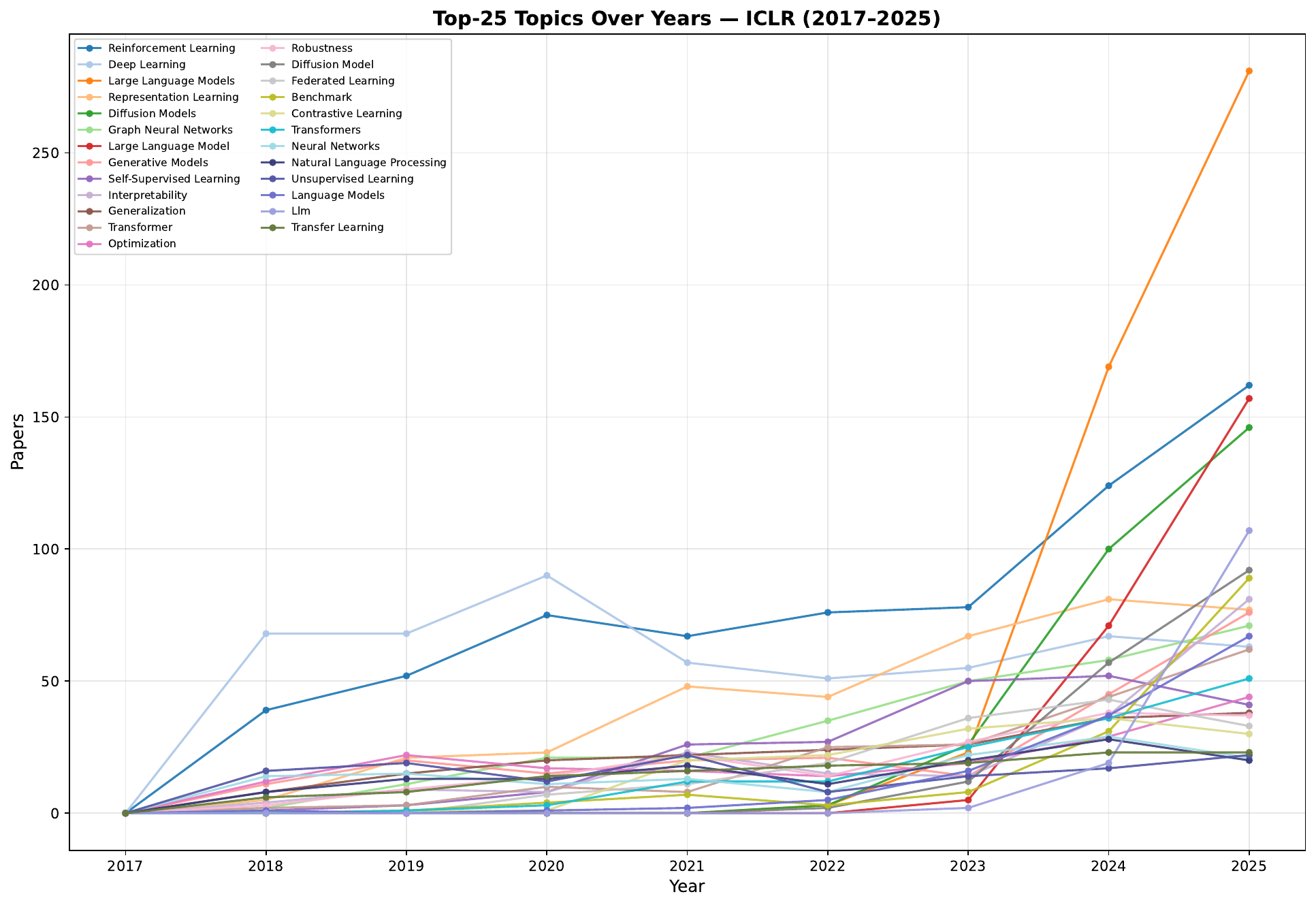}\\[2pt]
  {\small\textbf{(c) ICLR}}
\end{minipage}\hfill
\begin{minipage}[b]{0.49\linewidth}
  \centering
  \includegraphics[width=\linewidth]{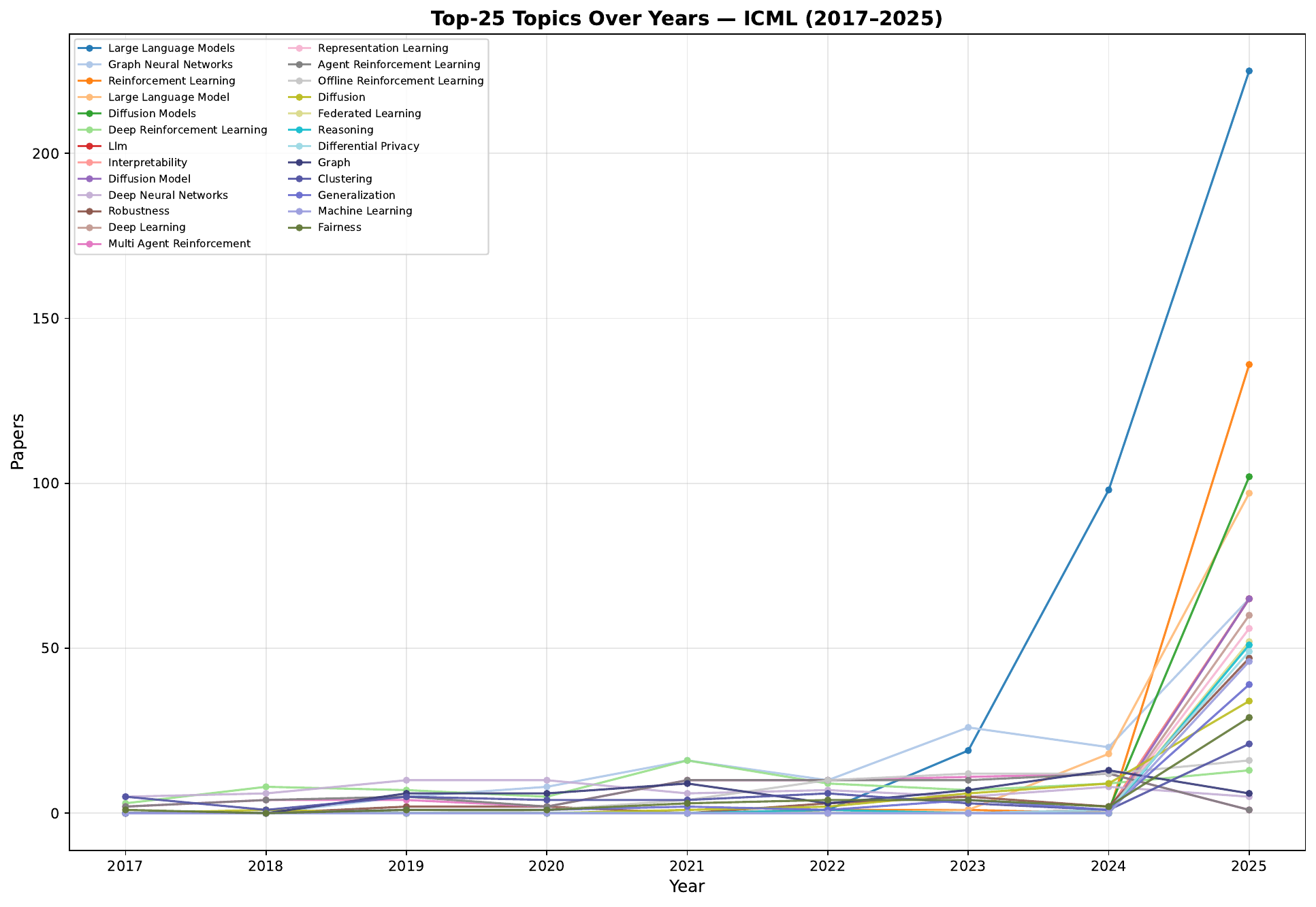}\\[2pt]
  {\small\textbf{(d) ICML}}
\end{minipage}

\vspace{0.8em}

\begin{minipage}[b]{0.49\linewidth}
  \centering
  \includegraphics[width=\linewidth]{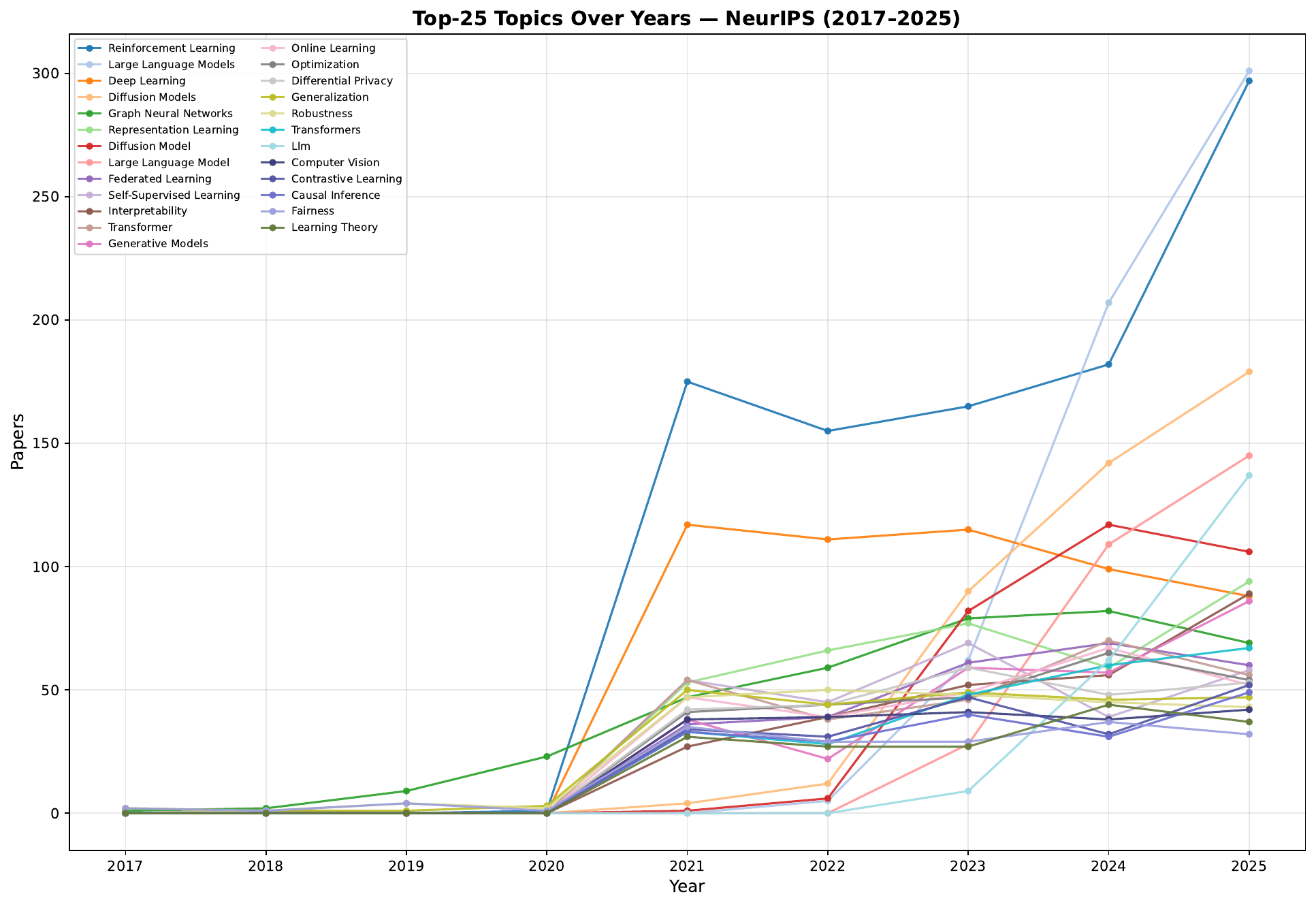}\\[2pt]
  {\small\textbf{(e) NeurIPS}}
\end{minipage}

\caption{Top-25 topics over years for each conference:
(a)~ACL, (b)~CVPR, (c)~ICLR, (d)~ICML, (e)~NeurIPS. Large language
models climb steeply from 2023 and overtake the long-standing leaders
(reinforcement learning, deep learning) by 2025 across ICLR, ICML, and
NeurIPS; at CVPR, vision-language and large language models rise while
3D object detection and neural radiance fields peak around 2023 and
decline. Established topics persist at high absolute volume even as
their relative share falls, most visibly at NeurIPS.}
\label{fig:top25_all}%
\label{fig:top25_acl}\label{fig:cvpr_topics}\label{fig:top25_iclr}%
\label{fig:top25_icml}\label{fig:top25_neurips}
\end{figure*}

NeurIPS (Fig.~\ref{fig:top25_neurips}), the largest venue in the
dataset, shows the transition at the greatest absolute scale.
\emph{Reinforcement learning} leads for most of the window and stays
near the top, but \emph{large language models} climbs from near-zero to
overtake it by 2025, with \emph{diffusion models} and
\emph{representation learning} rising into the upper band.
Because NeurIPS sustains more high-volume topics than any other venue,
it shows that the LLM transition coexists with, rather than erases,
established areas such as optimization, generalization, and graph
neural networks. Consolidation around dominant topics lowers the
relative share of the long tail without eliminating it: at a
sufficiently broad venue, established lines persist at high absolute
volume. NeurIPS is thus the clearest counterexample to reading
consolidation as wholesale replacement.

\end{document}